\documentclass[preprint,10pt]{elsarticle}

\usepackage{fontspec}
\IfFileExists{/home/ubuntu/workspace/DynaQ/fonts/times_new_roman.ttf}{%
  \setmainfont{TimesNewRoman}[
    Path = /home/ubuntu/workspace/DynaQ/fonts/,
    UprightFont = times_new_roman,
    BoldFont = times_new_roman_bold,
    ItalicFont = times_new_roman_italic,
    BoldItalicFont = times_new_roman_bold_italic,
    Extension = .ttf]
}{%
  \IfFileExists{fonts/times_new_roman.ttf}{%
    \setmainfont{TimesNewRoman}[
      Path = fonts/,
      UprightFont = times_new_roman,
      BoldFont = times_new_roman_bold,
      ItalicFont = times_new_roman_italic,
      BoldItalicFont = times_new_roman_bold_italic,
      Extension = .ttf]
  }{%
    \IfFontExistsTF{Times New Roman}{%
      \setmainfont{Times New Roman}
    }{%
      \setmainfont{TeX Gyre Termes}
    }
  }
}
\usepackage{newtxmath}          

\usepackage{geometry}
\usepackage{setspace}
\usepackage{etoolbox}
\AtBeginEnvironment{table}{\singlespacing}
\AtBeginEnvironment{algorithm}{\singlespacing}
\usepackage{afterpage}
\usepackage{caption}
\usepackage{amsmath,bm}
\usepackage{booktabs,multirow}
\usepackage{graphicx}
\usepackage{xcolor}
\usepackage[colorlinks=true,linkcolor=blue,citecolor=blue,urlcolor=blue]{hyperref}
\usepackage{algorithm}
\usepackage{algorithmic}

\graphicspath{{./}{./figs/}}

\journal{Pattern Recognition}

\begin{document}
\begin{frontmatter}

\title{ReHaSP: Relocating and harvesting structural priors for unsupervised semantic segmentation
of 3D point clouds}

\author[1]{Zhenghao Zhang}
\ead{zhangzhenghao25@nudt.edu.cn}
\author[1]{Xinjie Wang}
\ead{wangxinjie@nudt.edu.cn}
\author[2]{Yifan Zhang}
\ead{zhangyifan16c@nudt.edu.cn}
\author[1]{Wei Wang\corref{cor1}}
\cortext[cor1]{Corresponding author.}
\ead{wangwei_nudt@hotmail.com}
\author[1]{Jun Zhang}
\ead{zhangjun@nudt.edu.cn}
\author[3]{Hanyun Wang}
\ead{wanghanyun@mail.sysu.edu.cn}
\affiliation[1]{organization={College of Electronic Science and Technology, National University of
Defense Technology},
            city={Changsha},
            postcode={410073},
            country={China}}
\affiliation[2]{organization={Academy of Military Science},
            city={Beijing},
            postcode={100091},
            country={China}}
\affiliation[3]{organization={School of Electronics and Communication Engineering, Sun Yat-sen
University},
            city={Shenzhen},
            postcode={518107},
            country={China}}

\begin{abstract}
Unsupervised 3D semantic segmentation discovers semantic classes in point clouds without labels.
However, existing superpoint-based methods form pseudo-labels by clustering in feature space, without constraining superpoint merges, and predict each point independently at inference.
To fill this gap, we present ReHaSP, a framework that injects structural priors into pseudo-label generation and prediction.
At training time, spatially constrained clustering (SC) restricts merging to physically adjacent
superpoints, while multi-scale spectral analysis (MS) enriches superpoint descriptors across parts, objects, and scene regions.
This spatial constraint regularizes the enriched descriptors, and the enrichment makes the constraint highly effective.
At inference time, structure-aware inference (SI) fuses per-point with superpoint-level predictions through a label-free, training-free agreement confidence.
A mechanistic analysis indicates that training moves the residual errors away from spatially
remote fusions and concentrates them at superpoint boundaries, where inference harvests
them---a relocate-then-harvest pattern.
Extensive experiments on three benchmarks, cross-dataset transfer, and training-free
application of SI to other methods demonstrate that structural priors improve both
training-time pseudo-labels and inference-time predictions.
These gains hold across methods, transfer protocols, and all indoor benchmarks, where ReHaSP further outperforms published unsupervised methods.
Code will be released soon.
\end{abstract}


\end{frontmatter}
\suppressfloats[t]


\section{Introduction}
\label{sec:intro}

Enabling machines to discover the semantic composition of 3D environments automatically is crucial
for cutting-edge applications such as robotic navigation, augmented reality, and embodied
intelligence.
Over the past few years, fully supervised segmentation has advanced rapidly, with point-based
networks learning semantics on raw coordinates
\cite{qi2017pointnetplusplus} and sparse convolutions bringing efficient voxelized training
\cite{choy2019cvpr,graham2018sparseconv}.
These methods achieve strong accuracy but depend on large-scale point-level annotation.
Indeed, manually labeling real-world point clouds is prohibitively expensive, because the data
are unstructured and orderless, and a single indoor scan requires approximately 22.3 minutes of
expert labeling \cite{dai2017scannet}.
To reduce this burden, recent methods replace dense labels with weaker supervision, such as one
click per object \cite{liu2021onethingoneclick} or coarse hand-drawn strokes
\cite{unal2022scribble}.
These alternatives cut the labeling cost substantially yet still demand human effort that must be
repeated for every new dataset.
This motivates \emph{unsupervised} 3D semantic segmentation, the discovery of semantically
meaningful groups without point-level annotation for the target dataset.

\begin{figure*}[t]
	\centering
	\includegraphics[width=0.98\textwidth]{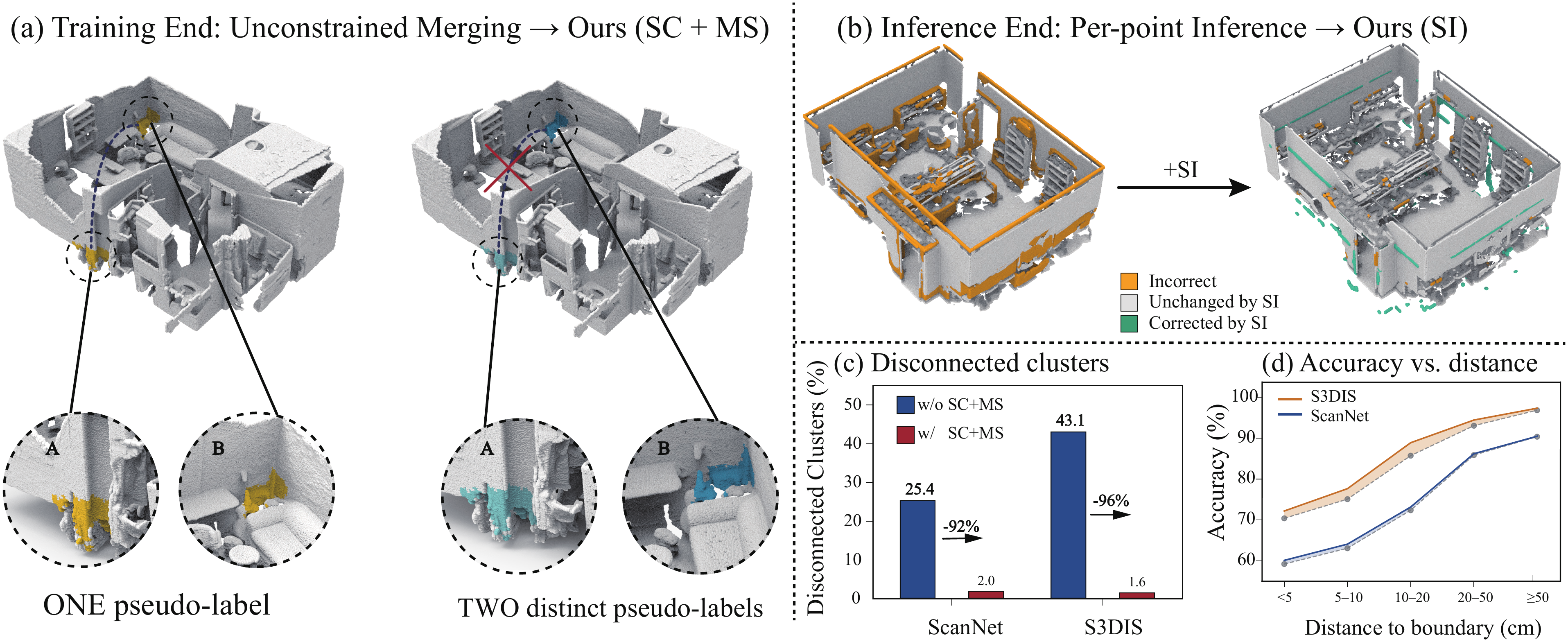}
	\caption{Structural priors at both ends of an unsupervised pipeline.
		(a)~Unconstrained merging assigns one pseudo-label to two spatially disconnected regions, a wall and a piece of furniture far apart.
		(b)~Per-point inference concentrates errors at superpoint boundaries, where SI corrects part of them and flips others.
		(c)~Spatially constrained clustering collapses the rate of disconnected clusters on both datasets (open, unconstrained KMeans; filled, SC).
		(d)~Accuracy by distance to the nearest superpoint boundary, per-point (dashed) and fused with SI (solid); the gain concentrates in the boundary band.}
	\label{fig:teaser}
\end{figure*}

In this regime, GrowSP \cite{Zhang_2023_CVPR} established the superpoint-growing paradigm from
geometry alone, without any pretrained features.
Later methods added cross-modal distillation and converged on a common four-stage procedure:
1)~distill 2D self-supervised features, DINO \cite{caron2021dino} and later DINOv2
\cite{oquab2024dinov2}, into a 3D backbone \cite{Chen_2023_ICCV};
2)~over-segment each scene into superpoints \cite{landrieu2018spg};
3)~cluster superpoint features into primitives \cite{Zhang_2025_CVPR};
and 4)~train the backbone on the resulting pseudo-labels.
Each generation sharpened one stage of the procedure, distillation \cite{Chen_2023_ICCV} importing
appearance priors into the descriptors, spectral grouping \cite{Zhang_2025_CVPR} reading global
patterns in the merge step, and recent refinements \cite{zhang2026growsppp,zhan2026pslcr} polishing
the resulting pseudo-labels.
A parallel line lifts SAM \cite{kirillov2023sam} or CLIP \cite{radford2021clip} masks into 3D
for zero-shot or label-free segmentation \cite{takmaz2023openmask3d,song2026pointgs}; these
methods are label-free like ours, but they transplant segmentations that were already produced in
2D, so the quality of those masks caps their accuracy.
We build within the superpoint family and ask what the procedure itself ignores.

The procedure nonetheless leaves both ends \emph{structurally blind}.
At the training end, pseudo-label generation is structurally blind in its merge rule and inference descriptor scale.
Superpoint merging is an unconstrained assignment in feature space.
Even LogoSP, whose affinity graph shapes the spectral embedding, allows any two superpoints to merge.
As shown in Fig.~\ref{fig:teaser}a, unconstrained merging assigns one shared pseudo-label to a
wall and a piece of furniture far apart.
Under the unconstrained KMeans merging of LogoSP \cite{Zhang_2025_CVPR}, remote cross-class fusions
of this kind recur across the indoor scenes we diagnose.
As shown in Fig.~\ref{fig:teaser}c, many merged clusters span spatially disconnected components.
Most of these fusions are dominated by a single class, so their damage is not to label purity but
to where the residual error sits.
Superpoint descriptors are pooled at a single fixed granularity.
Yet a chair back and an open floor differ by orders of magnitude, so a single pooling scale
cannot describe both.
At the inference end, the model predicts each point independently.
It thereby discards the superpoint structure on which the whole pipeline was built.
Supervised superpoint methods exploit this structure \cite{landrieu2018spg}, yet
unsupervised pipelines abandon it although they need it no less.
Label noise concentrates in the boundary bands between superpoints, where the pseudo-label error
rate is several times higher within centimeters of a boundary than deep inside superpoints.
As shown in Fig.~\ref{fig:teaser}b, the trained model inherits this error geography.
The open questions are therefore which superpoints may merge, at what scale each should be
described, and which superpoint predictions deserve trust at inference.

In this paper, we answer the first two with one coupled training-time mechanism and the third with
an inference-time prior. The three components are structural priors, rules built on
superpoints, their spatial relations, and the model's own predictions rather than on labels, and
they form the framework \textbf{ReHaSP}
(\emph{Re}locating and \emph{Ha}rvesting \emph{S}tructural \emph{P}riors).
At training time, the first two act as one mechanism. At inference time, the third acts alone.
1)~\emph{spatially constrained clustering} (SC), which answers the merge question by restricting
merging to physically adjacent superpoints. It nearly eliminates disconnected clusters, sharply
reduces remote cross-class contamination, and reads its constraint from scene geometry alone, as
the label-free setting requires;
2)~\emph{multi-scale spectral analysis} (MS), which answers the scale question by pooling superpoint
descriptors at several clustering granularities, so that chair backs and open floors are both
visible to the spectral stage;
and 3)~\emph{structure-aware inference} (SI), which answers the trust question by fusing per-point
predictions with superpoint-level predictions, weighted by a confidence estimated from
prediction agreement alone.
The SC--MS coupling is the key to the training stage. Enrichment without the adjacency rule
amplifies remote fusion, and the rule without enrichment does not help on its own.
The ablation quantifies this coupling.
Each prior alone degrades the procedure on both datasets, yet together they improve it.
For SI, the confidence is what makes fusion safe. Naively broadcasting superpoint votes can hurt,
and knowing when to trust the consensus separates fusion from noise amplification.
Its margins on each single dataset are small, but it needs no retraining and no labels, and as a
training-free transplant, it improves every base method on every dataset we tested, and every
cross-dataset transfer, turning negative only where superpoint consensus is unreliable.
The priors also cooperate across training and inference.
SC does not make the pseudo-labels purer overall; it moves their errors: wrong merges between
distant regions largely disappear, and what remains sits mostly at superpoint boundaries. The
reduction is large but not total, since merges can still travel far through chains of adjacent
superpoints.
This residual suits MS: pooling at several granularities sees past local boundary noise, where
single-scale pooling cannot.
SI's gains then concentrate at those same boundaries (Fig.~\ref{fig:teaser}d).
We call this the \emph{relocate-then-harvest} pattern, and a four-part mechanistic analysis
provides converging evidence for it.

Our main contributions can be summarized as follows:
\begin{itemize}
\item We propose \textbf{ReHaSP}, a framework that makes both pseudo-label generation and
prediction structurally aware, with SC and MS restructured as one coupled training-time mechanism
and SI fusing per-point with superpoint-level predictions at inference.
\item We propose a four-part mechanistic protocol that separates where pseudo-label errors sit from
how many there are, and use it to identify the relocate-then-harvest pattern.
\item We evaluate the framework on public indoor and outdoor benchmarks as well as bidirectional
cross-dataset transfers, where it sets or matches the best published unsupervised results on every
indoor benchmark while marking the negative boundary of SI outdoors.
\end{itemize}

\noindent A preliminary version of this work, comprising SC and MS, has been accepted at ICIG 2026
\cite{zhang2026mssp} and is available as a preprint.
This article extends it with structure-aware inference and its label-free confidence, the
four-part mechanistic protocol that identifies the relocate-then-harvest pattern, and a
validated applicability boundary including the outdoor failure mode and training-free transfers,
along with new ablations and sweeps.

\section{Related Work}
\label{sec:related}

\subsection{Unsupervised 3D semantic segmentation}
Fully supervised networks set the accuracy reference, from point-based backbones
\cite{qi2017pointnetplusplus} to sparse-convolutional encoders that voxelize scenes for efficient
3D convolution \cite{choy2019cvpr,graham2018sparseconv}, but their accuracy is bought with
point-level annotation, and weakly supervised variants only trade label density for accuracy
\cite{liu2021onethingoneclick,unal2022scribble}.
Early label-free methods replaced supervision with iterative clustering \cite{ji2019iic} or
mutual information across views
\cite{cho2021picie}, with limited semantic resolution.

GrowSP \cite{Zhang_2023_CVPR} established the superpoint-growing paradigm from geometry alone,
progressively merging superpoints with consistent local statistics. GrowSP++ \cite{zhang2026growsppp}
extends it with DINOv2 \cite{oquab2024dinov2} distillation as well as a refined growing schedule,
narrowing the gap to feature-based methods. Both merge by feature agreement alone, so neither rule
reads where the merged regions lie in the scene.
PointDC \cite{Chen_2023_ICCV} distills 2D self-supervised DINO \cite{caron2021dino} features
into 3D and clusters super-voxels in the distilled space, and LogoSP
\cite{Zhang_2025_CVPR} performs spectral analysis on a local--global superpoint graph, grouping
superpoints by their global patterns in the frequency domain, and reaches among the
strongest results in this family.
These methods sharpen what the descriptors contain, yet the grouping still happens in an
unconstrained feature space and at a single pooling granularity.
Training-time refinements polish pseudo-labels after they form rather than restructure how they
form. U3DS3 \cite{Liu_2024_WACV} refines them through spatial clustering and geometric invariance,
and P-SLCR \cite{zhan2026pslcr} filters unreliable supervision with a dynamic prototype library
and consistent reasoning.

A complementary line imports 2D foundation-model priors, SAM masks
\cite{kirillov2023sam} or
open-vocabulary features \cite{takmaz2023openmask3d}, to
segment 3D scenes zero-shot or, most recently, to discover classes without any labels
\cite{song2026pointgs}.
Its strongest instance, PointGS, lifts SAM masks via 3D Gaussian splatting, and small objects inherit the masks' sharpness, yielding strong per-class recall.
These methods share the label-free setting but transplant discrete 2D segmentation structure, so
their accuracy is capped by the quality of those masks rather than by the 3D grouping itself.

The present framework retains the distillation backbone but restructures what this lineage leaves
untouched, namely the merge rule, the descriptor granularity, and the prediction stage.
These questions are orthogonal to descriptor quality: no amount of feature enrichment determines where a merge may occur or which superpoint deserves trust during inference.
Where structure is exploited at prediction time, it is under supervision or with uniform
smoothing; we are not aware of an \emph{unsupervised} 3D segmenter that couples
spatially constrained pseudo-label formation with a confidence-gated superpoint fusion at
prediction time.

\subsection{Spectral clustering and multi-scale representation}
Spectral methods locate structure through the eigenspectrum of an affinity graph
\cite{shi2000ncut}, and deep spectral methods recently brought this machinery
to unsupervised 2D segmentation \cite{melaskyriazi2022deepspectral}, where feature eigenvectors
yield label-free segmentations of single images; superpoint graphs inherit the same machinery at
scene scale.
A descriptor that pools several granularities at once lets one spectral pass see all scales, which
fusing single-granularity partitionings after the fact cannot.
Hard connectivity likewise has a classical lineage.
Region growing and graph-based segmentation enforce that every output segment be spatially
connected \cite{felzenszwalb2004graph}, and connectivity-constrained agglomeration
\cite{scikit-learn} carries this prior into feature space.
There the constraint governs the final output partition under supervision or hand-tuned stopping
rules; in an unlabeled pipeline it must instead discipline an intermediate pseudo-label merge
whose correctness no labels can verify.
Classical normalized cuts instead wove proximity softly into the affinity itself, which helps
where proximity is informative but compromises where it stops being discriminative, as with
near-identical walls repeating across rooms.
Imposing hard connectivity on learned superpoint features inside an unlabeled pipeline remains
unexplored in this lineage.
Multi-scale representations, from feature pyramids \cite{lin2017fpn} and point-cloud encodings
\cite{qi2017pointnetplusplus} to recent unsupervised pipelines \cite{ren2025ucfseg}, enrich
descriptors across spatial extents.
In these works, the scales enter at the input encoder, whereas the superpoint descriptors that
feed the clustering graph have received no such multi-scale treatment.
The placement matters, since a single granularity forces the graph to compare chair backs against
whole rooms.

\subsection{Structural priors from superpixels to superpoints}
Superpixels and superpoints are the standard structural atoms of segmentation, in 2D
\cite{felzenszwalb2004graph} and 3D \cite{landrieu2018spg}: over-segmentation
follows object boundaries at low cost, leaving most label noise inside superpoint interiors.
This is precisely what an unlabeled pipeline can exploit: the boundaries come from geometry
rather than annotation, and the residual noise sits at a structurally characterized location
rather than being scattered.
Structural priors appear as pairwise smoothness in fully connected CRFs
\cite{krahenbuhl2011dense}, learnable feature-space graph convolution
\cite{Yang2022CRFConv}, connectivity-constrained clustering \cite{scikit-learn}, and
flexible neighborhood operators \cite{Floris2024Composite}.
At prediction time, exploiting structure also has a long history beyond superpoints.
CRF smoothing \cite{krahenbuhl2011dense} and graph-Laplacian regularization
\cite{zhou2004learning} refine predictions
from pairwise structure.
Supervised superpoint architectures classify directly at the superpoint level
\cite{landrieu2018spg}.
An unsupervised pipeline cannot copy the supervised designs, because no labels certify which
superpoints are trustworthy, and uniform smoothing is not a substitute, since it would crush the
minority points of mixed superpoints.
Agreement-weighted fusion also departs from test-time ensembling, which averages one predictor
over augmented inputs at uniform weight; SI instead fuses two structurally different predictors,
the per-point head and the superpoint consensus, and lets their agreement decide the weight of
each.
What is missing is a confidence estimate for superpoint trustworthiness based solely on prediction agreement.
\begin{figure*}[t]
	\centering
	\includegraphics[width=0.98\textwidth]{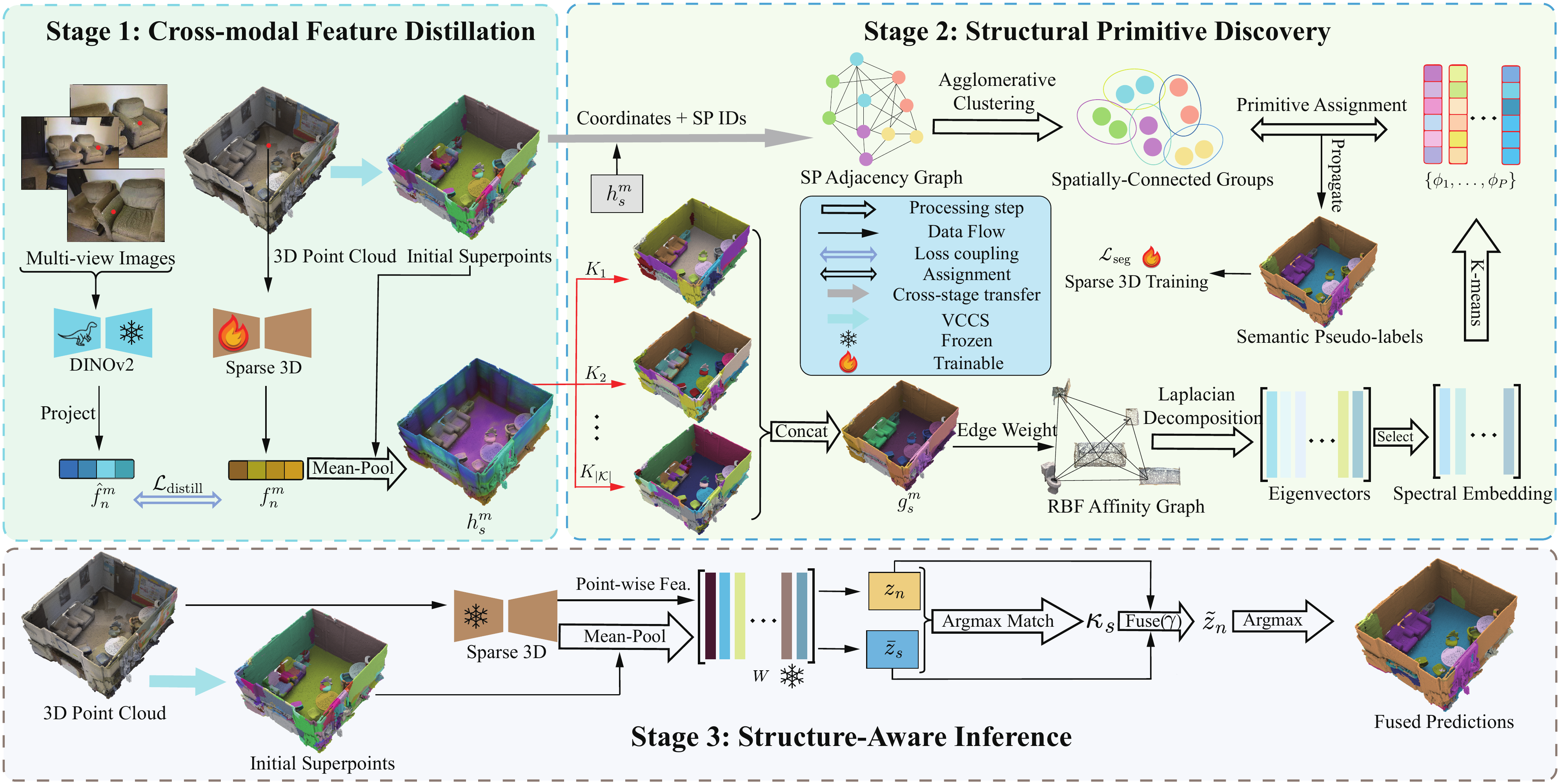}
	\caption{Overview of ReHaSP. Frozen 2D features are distilled into a sparse 3D backbone;
		superpoint descriptors fork into SC and MS and converge at primitive assignment to produce
		pseudo-labels; at inference, SI fuses per-point with superpoint-level predictions through a
		label-free agreement confidence.}
	\label{fig:pipeline}
\end{figure*}
Whether structural priors can act at pseudo-label formation and at prediction time, without any
labels, has remained open in this setting.

\section{The ReHaSP Framework}
\label{sec:method}

\subsection{Overview and problem formulation}
\label{sec:overview}
A training scene $\mathcal{P}^m$ is a point cloud of $N_m$ points with 3D coordinates and colors.
Alongside the points, each scene provides registered multi-view images and an initial
over-segmentation into $S_m$ superpoints $\{\mathcal{R}_s\}_{s=1}^{S_m}$ \cite{landrieu2018spg}.
The goal is to group all points into $P$ semantic primitives without any labels.
Class names enter only at evaluation time, through Hungarian alignment between primitives and
ground-truth categories.
The class count $P$ is the only prior, as in earlier unsupervised pipelines; we set $P$ per dataset to match the number of evaluated categories.

ReHaSP is a three-stage architecture that supplies structure at every point where an unsupervised
pipeline would otherwise discard it.
Stage 1 distills frozen 2D foundation appearance into a sparse 3D backbone
(Section~\ref{sec:prelim}).
Building on these features, Stage 2 discovers semantic primitives from superpoints under two coupled
priors, SC and MS, and produces the
pseudo-labels (Section~\ref{sec:discovery}).
Finally, Stage 3 decodes the trained backbone with superpoint-level consensus, weighted by an
unsupervised agreement confidence (Section~\ref{sec:si}).
Fig.~\ref{fig:pipeline} shows the overall flow, and Algorithm~\ref{alg:framework}
(\ref{app:alg}) details the complete procedure.

\subsection{Stage 1: Cross-modal feature distillation}
\label{sec:prelim}
Stage 1 distills 2D appearance into a 3D backbone, and every later stage builds on these features
(Fig.~\ref{fig:pipeline}, Stage~1).
A frozen DINOv2 vision transformer (ViT-S/14) \cite{oquab2024dinov2} first extracts per-pixel
features from the multi-view images.
Depth and camera parameters then project these 2D features onto the 3D points, following the
strongest superpoint pipelines \cite{Zhang_2025_CVPR,Chen_2023_ICCV}.
A point seen by several images receives the average of its projected features,
$\hat f_n^m = \frac{1}{|\mathcal{V}_n|}\sum_{v\in\mathcal{V}_n}\pi_v(n)$, where $\mathcal{V}_n$
collects the views observing point $n$ and $\pi_v(n)$ is the feature projected from view $v$, and a
point seen by none is skipped in the loss.
This projection is the framework's only external knowledge.
Each later stage is learned directly from the point cloud.
A sparse 3D backbone, Res16FPN18 \cite{choy2019cvpr}, takes the voxelized scene as input and
learns to reproduce the projected features $\hat f_n^m\in\mathbb{R}^D$ ($D{=}384$) under a
cosine-similarity objective.
On the indoor datasets, the backbone consumes only voxel coordinates and RGB colors, and
coordinates serve voxelization, not input features; the nuScenes backbone instead consumes xyz
(Table~\ref{tab:hyper}).
\begin{equation}
	\mathcal{L}_{\text{distill}} = \frac{1}{N_m}\sum_{n=1}^{N_m}\bigl(1-\cos(f_n^m,\hat f_n^m)\bigr).
	\label{eq:distill}
\end{equation}
Here $f_n^m$ denotes the backbone output at point $n$, $\hat f_n^m$ the projected target, and
$N_m$ the number of points of scene $m$ seen by at least one view.
The cosine compares feature directions rather than magnitudes, matching how Stage 2 measures
similarity.
Point features are average-pooled within each initial superpoint:
\begin{equation}
	h_s^m = \frac{1}{|\mathcal{R}_s|}\sum_{n\in\mathcal{R}_s} f_n^m \in \mathbb{R}^D .
	\label{eq:spfeat}
\end{equation}

\subsection{Stage 2: Structural primitive discovery}
\label{sec:discovery}
This stage answers two design questions (Fig.~\ref{fig:pipeline}, Stage~2): which superpoints may
merge into one pseudo-label---unconstrained merging is spatially blind, hard adjacency alone
sacrifices feature-driven grouping---and at what granularity descriptors should be built, since a
single granularity cannot represent both object parts and whole objects.
The two answers, SC and MS, are one
mechanism in a \emph{regularize-then-enrich} arrangement: SC supplies the spatial coherence that
MS's enriched descriptors require, and MS provides the representation that makes merging worthwhile;
neither half alone improves over the baseline (Section~\ref{sec:ablation}).
The fork converges again at primitive assignment.

\subsubsection{Spatially constrained clustering}
\label{sec:sc}
The obvious rule merges superpoints solely by feature similarity. The KMeans in LogoSP's spectral
embedding does exactly this. Its spectral analysis captures global feature-space structure, but the
merge step itself can still fuse spatially non-adjacent superpoints into one pseudo-label.
The remedy is to make scene adjacency part of the rule itself: we construct a superpoint adjacency
graph $\mathcal{G}^m=(\mathcal{V},\mathcal{E})$, $\mathcal{V}=\{1,\dots,S_m\}$, with an edge
whenever any voxel pair is 6-adjacent:
\begin{equation}
	\mathcal{E} = \bigl\{(i,j)\;\big|\;\exists\,v_i\in\mathcal{V}(\mathcal{R}_i),\,
	v_j\in\mathcal{V}(\mathcal{R}_j):\ \|v_i-v_j\|_1=1\bigr\},
	\label{eq:adj}
\end{equation}
where $\mathcal{V}(\mathcal{R})$ collects the voxels of superpoint $\mathcal{R}$ and $v$ denotes a
voxel coordinate.
We then replace KMeans with connectivity-constrained agglomerative clustering \cite{scikit-learn},
merging the $S_m$ initial superpoints into $C$ spatially-connected groups on which primitive
assignment operates.
The graph itself is purely geometric: an edge records voxel contact, nothing about feature
similarity.
The merge is greedy Ward agglomeration on the L2-normalized descriptors.
Each step fuses the adjacent pair whose union least increases the within-group feature variance,
\begin{equation}
	(i^\ast, j^\ast) = \arg\min_{(i,j)\in\mathcal{E}} \Delta(i,j),\qquad
	\Delta(i,j) = \frac{n_i n_j}{n_i + n_j}\lVert \mu_i - \mu_j\rVert_2^2,
	\label{eq:ward}
\end{equation}
where $n_i$ and $\mu_i$ are the size and mean descriptor of group $i$, so every merge respects
adjacency.
The result is a three-level granularity ladder, from the $S_m$ initial superpoints, through the $C$
consolidated groups, to the $P$ primitives.
One caveat is honest bookkeeping: isolated islands in the adjacency graph can still share a cluster
after graph completion, so SC reduces disconnected clusters to a small residue rather than zero
(Section~\ref{sec:mechanism}).

\subsubsection{Multi-scale spectral analysis}
\label{sec:ms}
Superpoint methods \cite{Zhang_2025_CVPR,Zhang_2023_CVPR,Chen_2023_ICCV} pool descriptors at one
fixed granularity, yet semantic parts and whole objects live at very different spatial scales, so no
single granularity can see them all.
The insight is to make the descriptor multi-granular by construction: for
$\mathcal{K}=\{K_1,\dots,K_{|\mathcal{K}|}\}$ , we run KMeans on the
L2-normalized descriptors $\hat h_s^m = h_s^m/\lVert h_s^m\rVert_2$ at each granularity $K$, then
re-pool each cluster centroid in the original feature space,
\begin{equation}
	\mu_k^{(K)} = \tfrac{1}{|\mathcal{C}_k^{(K)}|}\sum_{s\in\mathcal{C}_k^{(K)}} h_s^m,\qquad
	g_s^{m,(K)} = \mu_{\pi_K(s)}^{(K)} / \lVert \mu_{\pi_K(s)}^{(K)} \rVert_2,
	\label{eq:ms_centroid}
\end{equation}
and concatenate $g_s^m =
[\,g_s^{m,(K_1)};\dots;g_s^{m,(K_{|\mathcal{K}|})}\,]\in\mathbb{R}^{D\cdot|\mathcal{K}|}$,
where $\mathcal{C}_k^{(K)}$ is the $k$-th cluster of the granularity-$K$ partition within the
scene (scene index $m$ omitted) and $\pi_K(s)$ maps superpoint $s$ to its cluster.
The finest $K$ separates object parts, the coarsest approaches whole objects, and the descriptor
carries all three views at once.
Each KMeans runs within one scene, while the spectral stage sees all scenes at once, making the
descriptors local but the grouping global.
With $|\mathcal{K}|$ granularities the descriptor has $D\cdot|\mathcal{K}|$ dimensions, and the
spectral pass dominates memory; \ref{app:cost} quantifies the cost.
The spectral stage consumes initial superpoints, not the $C$ merged ones, and pools them
across all training scenes, capped by per-scene stratified sampling (Section~\ref{sec:setup}).

\subsubsection{Spectral primitive discovery}
\label{sec:spectral}
The pooled descriptors mix multiple granularities, so we let the spectrum rather than any single
feature axis decide the grouping.
An RBF affinity matrix is built on the pooled set $\{g_s\}$:
\begin{equation}
	A_{ij} = \exp\!\Bigl(-\frac{\lVert g_i - g_j\rVert_2^2}{2\sigma^2}\Bigr),
	\label{eq:rbf}
\end{equation}
where $\sigma$ is the median pairwise distance among $\{g_s\}$.
The normalized Laplacian is
\begin{equation}
	L = I - D^{-\frac{1}{2}} A D^{-\frac{1}{2}},
	\label{eq:laplacian}
\end{equation}
where $D$ is the degree matrix of $A$.
Its eigenvectors with the smallest eigenvalues---the leading eigenvectors of the
normalized affinity $D^{-1/2}AD^{-1/2}$, after dropping the trivial constant
one---give every superpoint a spectral embedding $u_s\in\mathbb{R}^{d}$.
Eigenvectors are kept greedily until they cover $99\%$ of the descriptors' projection energy,
retaining about $50$ columns.
These eigenvectors act as global patterns shared across scenes.
We keep the spectral stage of LogoSP \cite{Zhang_2025_CVPR} because its global patterns absorb the
redundancy that the concatenated, multi-granularity descriptor introduces.
A final KMeans on the embedding partitions the sampled superpoints into $P$ primitives
$\{\phi_1,\dots,\phi_P\}$.

Both halves converge at primitive assignment.
Each primitive center is the mean of its member superpoints' single-scale features $h_s^m$, and the
$P$ normalized centers form a linear classifier $W = [\,w_1,\dots,w_P\,]\in\mathbb{R}^{D\times P}$
that shares the backbone's output space.
Each merged group takes the label of its nearest center, and the label propagates to every point
inside,
\begin{equation}
	w_p = \frac{\bar\mu_p}{\lVert\bar\mu_p\rVert_2},\quad
	\bar\mu_p = \frac{1}{|\varphi_p|}\sum_{s\in\varphi_p} h_s^m,\qquad
	\hat y_n = \arg\max_p\, \cos\bigl(\mu_{g(n)}, w_p\bigr),
	\label{eq:assign}
\end{equation}
where $\varphi_p$ collects the superpoints of primitive $p$, and $g(n)$ and $\mu_{g(n)}$ denote the
merged group containing point $n$ and its mean descriptor.
The pseudo-labels then drive a second training pass of the backbone.
The classifier $W$ stays frozen, and the backbone is updated by minimizing the cross-entropy
between each pseudo-label and the temperature-scaled cosine logits,
\begin{equation}
	\mathcal{L}_{\text{seg}} = \frac{1}{N_m}\sum_{n=1}^{N_m}
	\ell_{\mathrm{CE}}\bigl(\tau\cos(f_n^m, W),\, \hat y_n^m\bigr),
	\label{eq:ce}
\end{equation}
where $\tau$ is a scalar temperature, $\hat y_n^m$ is the pseudo-label of point $n$, and
$\cos(f_n^m, W)$ stacks the cosine similarities of the point feature against the $P$ columns of
$W$.
This pass turns the distilled appearance features into features that separate the discovered
primitives.
The pseudo-labels are fixed after the initial clustering, so the supervision never drifts.
Because the spectral path consumes initial-superpoint features only, SC modifies neither the
spectrum nor the primitives: its entire effect is on merge composition and pseudo-label granularity,
the property our mechanistic analysis exploits to isolate the priors' effect (Section~\ref{sec:mechanism}).

\begin{figure*}[t]
	\centering
	\includegraphics[width=0.98\textwidth]{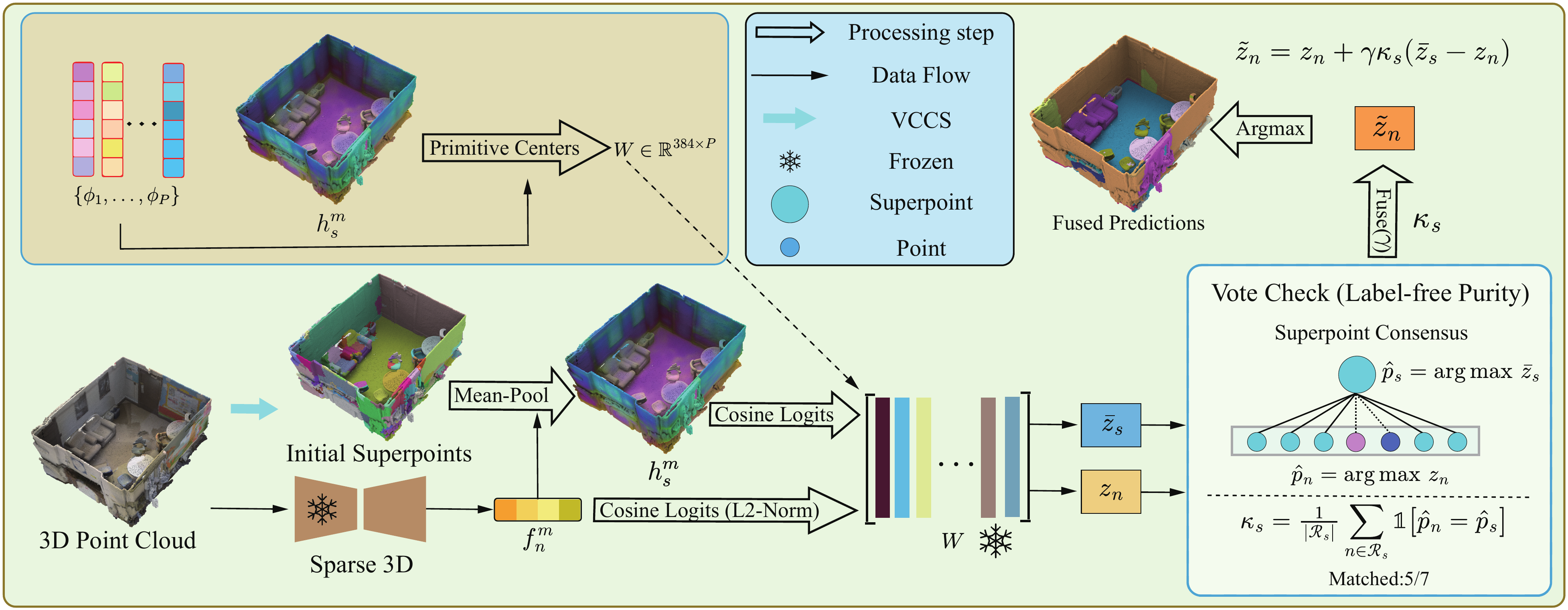}
	\caption{Structure-aware inference in detail. The shared backbone produces per-point features
		that fork into two paths: per-point logits $z_n$ and superpoint consensus logits $\bar z_s$
		(mean-pooled within each initial superpoint, then projected through the same frozen classifier $W$).
		A vote check compares the two predictions to yield the label-free agreement confidence $\kappa_s$,
		which gates the fusion of Eq.~\ref{eq:fusion}.}
	\label{fig:scvis}
\end{figure*}

\subsection{Stage 3: Structure-aware inference}
\label{sec:si}
After training, the standard protocol predicts each point independently, discarding superpoint
structure (Fig.~\ref{fig:pipeline}, Stage~3).
The obvious fix, broadcasting each superpoint's consensus to all of its points, crushes the
minority points of mixed superpoints and measurably hurts on one of our datasets
(Section~\ref{sec:sianalysis}).
SI instead scores each point twice, once by the backbone and once by its superpoint, and fuses the
two scores with a confidence that knows how trustworthy the superpoint is
(Fig.~\ref{fig:scvis} details this procedure).

\paragraph{Dual scoring}
The point logits $z_n\in\mathbb{R}^{P}$ and the superpoint logits $\bar z_s\in\mathbb{R}^{P}$ of the
superpoint $\mathcal{R}_s$ containing $n$ are both cosine similarities against the L2-normalized
columns of $W$, consistent with the distillation objective (Eq.~\ref{eq:distill}),
\begin{equation}
	z_n = \cos\bigl(f_n^m, W\bigr),\qquad
	\bar z_s = \cos\Bigl(\tfrac{\bar f_s}{\lVert \bar f_s\rVert_2},\, W\Bigr),\quad
	\bar f_s = \frac{1}{|\mathcal{R}_s|}\sum_{n\in\mathcal{R}_s} f_n^m,
	\label{eq:dual}
\end{equation}
where the cosine stacking follows Eq.~\ref{eq:ce} and $\bar f_s$ equals $h_s^m$ from
Eq.~\ref{eq:spfeat}.
The consensus uses the initial partition, which is purely geometric and therefore available at
inference.

\paragraph{Label-free agreement confidence}
We estimate an unsupervised confidence per superpoint from prediction agreement alone:
\begin{equation}
	\kappa_s = \frac{1}{|\mathcal{R}_s|}\sum_{n\in\mathcal{R}_s}\mathbf{1}\bigl[\arg\max z_n =
	\arg\max \bar z_s\bigr].
	\label{eq:conf}
\end{equation}
Here $\mathbf{1}[\cdot]$ denotes the indicator function, which equals $1$ when its argument is
true and $0$ otherwise.
This is the fraction of points that already agree with their superpoint, with no labels involved.

\paragraph{Confidence-weighted fusion}
The fused logits are
\begin{equation}
	\tilde z_n = z_n + \gamma\,\kappa_{s(n)}\bigl(\bar z_{s(n)} - z_n\bigr),
	\label{eq:fusion}
\end{equation}
with a single scalar $\gamma$.
The product $\gamma\kappa_s$ acts as a per-superpoint interpolation weight.
At zero, the per-point prediction is untouched; at one, the superpoint consensus takes over
completely.
Broadcasting replaces every prediction, whereas fusion interpolates between the two per
superpoint, by how much trust the superpoint has earned.

\paragraph{Behavior and cost}
When a superpoint is pure, with $\kappa_s\!\approx\!1$, its boundary points are pulled toward the
superpoint consensus, smoothing noise exactly where per-point evidence is weakest.
When it is mixed, with $\kappa_s$ low, fusion is gentle and informative per-point evidence survives.
On a homogeneous wall superpoint, the few stray predictions snap to the wall class; on a superpoint
straddling a door frame, both classes keep their own evidence.
Fusion also operates on logits rather than features, so it composes with any backbone and never
retrains $W$.
SI requires no training, no labels, and one hyperparameter, and adds only one superpoint pooling
pass per scene at negligible cost (\ref{app:cost}).
It applies to superpoint-based unsupervised segmenters whose superpoint consensus is reliable, a
boundary Section~\ref{sec:sianalysis} quantifies.

\section{Experiments}
\label{sec:experiments}

\subsection{Experimental settings and datasets}
\label{sec:setup}
\textbf{Datasets.} S3DIS \cite{armeni20163d} contains six indoor areas with 13 classes, and we set
$P{=}12$.
We evaluate the Area-5 protocol, which trains on the other five areas, and 6-fold
cross-validation, which holds out each area in turn.
ScanNet V2 \cite{dai2017scannet} contains 1{,}201 training and 312 validation single-room scenes
with 20 classes, voxelized at 0.05\,m, and we set $P{=}20$.
nuScenes \cite{caesar2020nuscenes} provides outdoor LiDAR scans of autonomous-driving scenes with
16 annotated categories and 28{,}130 training scans.
Our nuScenes models are trained on the first 400 scans in canonical order, evaluated on the full
6{,}018-scan validation split, against baselines trained on the full set, so the comparison is
conservative by a factor of $70$ in training data.

\textbf{Implementation.} The Res16FPN18 backbone is distilled for 300 epochs on ScanNet and 700 on
S3DIS, and on nuScenes we adopt the released LogoSP distillation checkpoint, likewise trained for
300 epochs.
The scale set is $\mathcal{K}{=}\{80,40,20\}$, the merged-group count is $C{=}40$ on ScanNet and
$80$ on S3DIS, and the spectral sample cap is 30{,}000 superpoints with per-scene stratified
sampling.
Outdoor nuScenes scenes yield far fewer superpoints per scene, so the scales reduce to
$\{20,10,5\}$ and the merged-group count to $C{=}20$ there.
Cross-entropy training runs 200 epochs on ScanNet, 100 on S3DIS, and 50 on nuScenes, at
temperature 5 on ScanNet and 3 on S3DIS and nuScenes.
SI uses $\gamma{=}0.4$ on ScanNet and nuScenes and $0.8$ on S3DIS unless stated.
For clarity of attribution, rows labeled \emph{MSSP} \cite{zhang2026mssp} denote our conference
version, i.e., the training-time components MS and SC alone, whereas \emph{ReHaSP} denotes the
full framework, which adds inference-time SI on the same backbone.
Table~\ref{tab:hyper} in \ref{app:hyper} lists all hyperparameters.

\textbf{Metrics.} Following common practice, we report overall accuracy, mean class accuracy, and
mean intersection over union, with mIoU as the primary metric.
mIoU compounds the point-level and class-equal perspectives and serves as the primary criterion
throughout.
Because discovered primitives carry no semantic meaning, one Hungarian assignment per split aligns
them with ground-truth classes.
The assignment is computed once per evaluation split from the aggregated confusion matrix,
identically for every method.
Variance, where reported, reflects independent training seeds rather than scene sampling.
\begin{figure*}[t]
	\centering
	\includegraphics[width=0.98\textwidth]{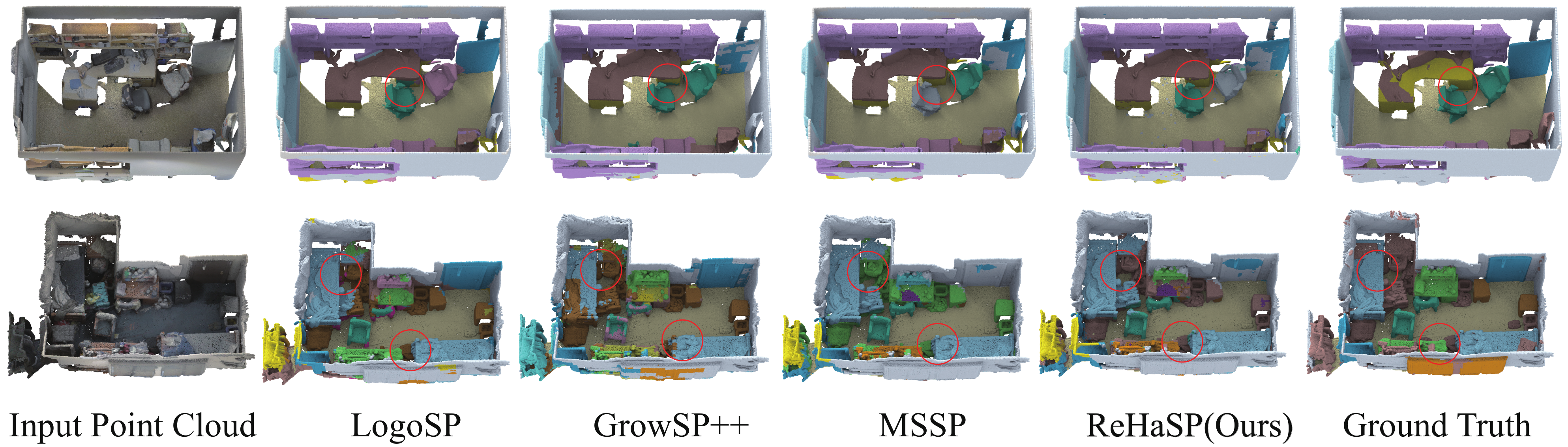}
	\caption{Qualitative comparison on indoor datasets: S3DIS Area 5 (first row) and ScanNet (second
		row). ReHaSP produces coherent primitives with cleaner object boundaries than its
		training-time variant (MSSP).}
	\label{fig:qual}
\end{figure*}
\subsection{Evaluation on benchmarks}
\label{sec:sota}
Across the benchmarks below, the training-time priors improve every dataset, and SI without
retraining improves every method it touches on every indoor dataset
(Table~\ref{tab:universal}); Section~\ref{sec:siboundary} marks where it does not.
The two indoor benchmarks stress remote fusion and boundary noise under different room
statistics, and the outdoor benchmark stresses superpoint purity.

\subsubsection{Evaluation on S3DIS}

\textbf{Dataset.}
S3DIS rooms include RGB-D images aligned to the point clouds, enabling our 2D-to-3D distillation.
Following common practice, the heterogeneous ``clutter'' class is excluded from evaluation,
leaving 12 evaluated classes.

\textbf{Results.}
As shown in Table~\ref{tab:s3dis} and Fig.~\ref{fig:qual}, ReHaSP reaches \textbf{50.7} mIoU and 83.9 OA on
Area 5, the best unsupervised mIoU overall and the best OA among all methods from other groups,
surpassing the strongest
published result, PointGS \cite{song2026pointgs} at $49.3$, by $1.4$.
Both priors contribute: the training-time components lift the score from LogoSP's $46.5$ to
MSSP's $49.2$, and SI adds another $1.5$ on top.
The $50.7$ mIoU also surpasses the supervised PointNet++ \cite{qi2017pointnetplusplus} ($50.1$; OA
$83.9$ against $77.5$).
Sparse-convolutional supervision remains clearly ahead at $60.8$ mIoU.
Under the 6-fold protocol, also reported in Table~\ref{tab:s3dis}, ReHaSP reaches $49.3$ mIoU with the best
unsupervised OA of $81.2$, surpassing P-SLCR's $47.5$ mIoU by $1.8$.
The training-time components alone already reach $48.5$, also above every published unsupervised
result, and SI adds $+0.8$ on the same checkpoints ($+0.81 \pm 0.66$ across the six folds,
paired $t$-test $p{=}0.030$, positive on five of six).

\textbf{Analysis.}
Area 5 concentrates large open spaces, where unconstrained merging fuses disconnected structures
most aggressively, so the priors have abundant remote fusion to remove---mostly benign
individually (Section~\ref{sec:mechanism}), yet their removal reshapes the label field SI
harvests.
Section~\ref{sec:mechanism} measures consistently higher remote contamination on S3DIS than on
ScanNet.
The 6-fold protocol instead averages smaller rooms across folds, which may dilute boundary-specific gains.
Area 5 thus exposes how much remote fusion there is to remove, while the 6-fold average tests
whether the gains survive less favorable scene statistics.
The per-class profile (\ref{app:perclass}) is uneven rather than uniform: because Hungarian
matching assigns primitives to classes in one global step, a single sofa-like primitive can take
over the entire sofa class, lifting its IoU from $21.9$ to $64.8$, while most other classes
decline slightly, with chair down $7.7$ and floor down $4.2$.
The same global matching explains PointGS's \cite{song2026pointgs} mAcc lead, which comes from
mask sharpness on small objects, yet its overall accuracy still trails ours, $76.6$ against
$83.9$; SI, by contrast, adds small positive margins almost uniformly.

\begin{table}[t]
	\centering
	\caption{Results on S3DIS Area 5 and 6-fold cross-validation (OA/mAcc/mIoU, \%); best unsupervised
		per protocol in \textbf{bold}. The two 6-fold rows share one set of checkpoints, so
		their difference isolates SI on the same backbone.}
	\label{tab:s3dis}
	\setlength{\tabcolsep}{3pt}
	\begin{tabular}{l ccc ccc}
		\toprule
		& \multicolumn{3}{c}{Area 5} & \multicolumn{3}{c}{6-fold} \\
		\cmidrule(lr){2-4}\cmidrule(lr){5-7}
		Method & OA & mAcc & mIoU & OA & mAcc & mIoU \\
		\midrule
		\multicolumn{7}{l}{\textit{Supervised}} \\
		PointNet++ \cite{qi2017pointnetplusplus} & 77.5 & 62.6 & 50.1 & 77.1 & 74.1 & 55.1 \\
		SparseConv \cite{graham2018sparseconv}   & 88.4 & 69.2 & 60.8 & 89.4 & 78.1 & 69.2 \\
		\midrule
		\multicolumn{7}{l}{\textit{Unsupervised}} \\
		GrowSP$^\dagger$ \cite{Zhang_2023_CVPR} & 78.4 & 57.2 & 44.5 & 76.0 & 59.4 & 44.6 \\
		PointDC \cite{Chen_2023_ICCV}            & 54.1 & --   & 22.6 & 55.7 & 37.7 & 26.0 \\
		PointDC-DINOv2 \cite{Chen_2023_ICCV}     & 75.7 & 48.7 & 40.2 & 74.4 & 51.5 & 41.3 \\
		U3DS3$^\dagger$ \cite{Liu_2024_WACV}     & 75.5 & 55.8 & 42.8 & --   & --   & -- \\
		IIC \cite{ji2019iic}                     & 28.5 & 12.5 & 6.4  & 32.8 & 14.7 & 8.5 \\
		PiCIE \cite{cho2021picie}                & 61.6 & 25.8 & 17.9 & 46.4 & 28.1 & 17.8 \\
		LogoSP \cite{Zhang_2025_CVPR}            & 82.8 & 55.9 & 46.5 & 79.2 & 58.0 & 46.3 \\
		GrowSP++ \cite{zhang2026growsppp}         & 78.7 & 60.1 & 46.6 & 77.6 & 59.9 & 47.1 \\
		P-SLCR \cite{zhan2026pslcr}               & 79.7 & 57.2 & 47.1 & 79.0 & \textbf{62.2} & 47.5 \\
		PointGS$^\S$ \cite{song2026pointgs}       & 76.6 & \textbf{66.1} & 49.3 & --   & --   & -- \\
		MSSP \cite{zhang2026mssp}                 & \textbf{84.2} & 56.8 & 49.2 & 80.2 & 61.1 & 48.5 \\
		\textbf{ReHaSP (Ours)}                   & 83.9 & 57.5 & \textbf{50.7} & \textbf{81.2} & 61.4 & \textbf{49.3} \\
		\bottomrule
	\end{tabular}
	\par\vspace{3pt}\parbox{\linewidth}{\footnotesize
		$^\dagger$No pretrained 2D features; $^\S$SAM-lifted 2D masks; `--' = not reported.}
\end{table}
\afterpage{\begin{table}[t] 	\centering 	\caption{Results on the ScanNet validation split (OA/mAcc/mIoU, \%) and on the official online 		benchmark (hidden test set; mIoU, \%). Best unsupervised results per protocol in \textbf{bold}.} 	\label{tab:scannet} 	\label{tab:scannet_online} 	\setlength{\tabcolsep}{4pt} 	\begin{tabular}{l ccc c} 		\toprule 		& \multicolumn{3}{c}{Validation split} & Online \\ 		\cmidrule(lr){2-4}\cmidrule(lr){5-5} 		Method & OA & mAcc & mIoU & mIoU \\ 		\midrule 		\multicolumn{5}{l}{\textit{Supervised}} \\ 		PointNet++ \cite{qi2017pointnetplusplus} & -- & -- & -- & 33.9 \\ 		SparseConv \cite{graham2018sparseconv}   & -- & -- & -- & 72.5 \\ 		\midrule 		\multicolumn{5}{l}{\textit{Unsupervised}} \\ 		GrowSP$^\dagger$ \cite{Zhang_2023_CVPR} & 57.3 & 44.2 & 25.4 & 26.9 \\ 		PointDC \cite{Chen_2023_ICCV}            & 63.7 & --   & 25.7 & 22.9 \\ 		PointDC-DINOv2 \cite{Chen_2023_ICCV}     & 64.7 & 45.0 & 29.6 & -- \\ 		U3DS3$^\dagger$ \cite{Liu_2024_WACV}     & 60.1 & 46.8 & 27.3 & -- \\ 		IIC \cite{ji2019iic}                     & 27.7 & 6.1 & 2.9 & -- \\ 		PiCIE \cite{cho2021picie}                & 20.4 & 16.5 & 7.6 & -- \\ 		LogoSP \cite{Zhang_2025_CVPR}            & 64.7 & 50.8 & 35.8 & 32.7 \\ 		GrowSP++ \cite{zhang2026growsppp}         & 70.5 & 48.8 & 33.2 & 32.3 \\ 		P-SLCR \cite{zhan2026pslcr}               & 61.4 & 49.0 & 29.0 & -- \\ 		PointGS$^\S$ \cite{song2026pointgs}       & --   & --   & \textbf{36.7} & -- \\ 		MSSP \cite{zhang2026mssp}                 & 70.7 & \textbf{51.3} & 36.2 & 34.7 \\ 		\textbf{ReHaSP (Ours)}                   & \textbf{71.4} & \textbf{51.3} & 36.6 & \textbf{35.1} \\ 		\bottomrule 	\end{tabular} 	\par\vspace{3pt}\parbox{\linewidth}{\footnotesize 		$^\dagger$No pretrained 2D features; $^\S$SAM-lifted 2D masks (OA/mAcc not reported); 		`--' = not submitted or not reported.} \end{table}}
\subsubsection{Evaluation on ScanNet}
\textbf{Dataset.}
ScanNet scans carry RGB-D alignment; undefined background points are excluded from clustering,
loss, and metrics.
Unlike S3DIS areas, ScanNet scenes are single rooms with structurally repetitive layouts such as
walls, floors, cabinets, and chairs.
Unconstrained merging therefore produces less remote contamination, $13.8\%$ per
Section~\ref{sec:mechanism}, but denser superpoint boundaries.

\textbf{Results.}
As shown in Table~\ref{tab:scannet}, ReHaSP reaches \textbf{36.6} mIoU on the validation split,
surpassing LogoSP by $0.8$ and P-SLCR by $7.6$.
This matches the strongest published result, PointGS's $36.7$.
OA improves to $\textbf{71.4}$, a gain of $6.7$ over LogoSP and the largest among unsupervised
methods.
On the official online benchmark of Table~\ref{tab:scannet_online}, the full framework reaches
\textbf{35.1} mIoU, leading LogoSP by $2.4$.
PointNet++ stands at $33.9$ there; the only unsupervised methods above it are ours (MSSP $34.7$,
ReHaSP $35.1$).

\textbf{Analysis.}
The smaller margin than on S3DIS is expected, since less remote merging leaves the
relocation--harvest loop less to fix.
Both components still contribute positively, and the largest OA gain among unsupervised methods
indicates that the priors act on the large coherent structures of these single-room scenes.
The val$\to$test drop is also smaller than LogoSP's ($-1.5$ vs.\ $-3.1$), consistent with
transfer rather than overfitting the validation split.
The SI margin is not seed luck: $+0.37 \pm 0.08$ mIoU across six independently trained seeds
(paired $t$-test $p{<}10^{-4}$, six of six positive), stable while absolute seed performance
itself varies by more than $2$ mIoU, and positive on $85\%$ of individual validation scenes
(per-scene bootstrap $95\%$ CI: $68$--$89\%$ of scenes).

\subsubsection{Evaluation on nuScenes}
\textbf{Dataset.}
Outdoor scenes differ from indoor ones in two structural respects: LiDAR sampling makes superpoints
small and point-wise impure, roughly $57$ points per superpoint against ${\sim}10^3$ indoors, and
the category distribution is extremely dominated by road-like surfaces.

\begin{figure*}[t]
	\centering
	\includegraphics[width=0.98\textwidth]{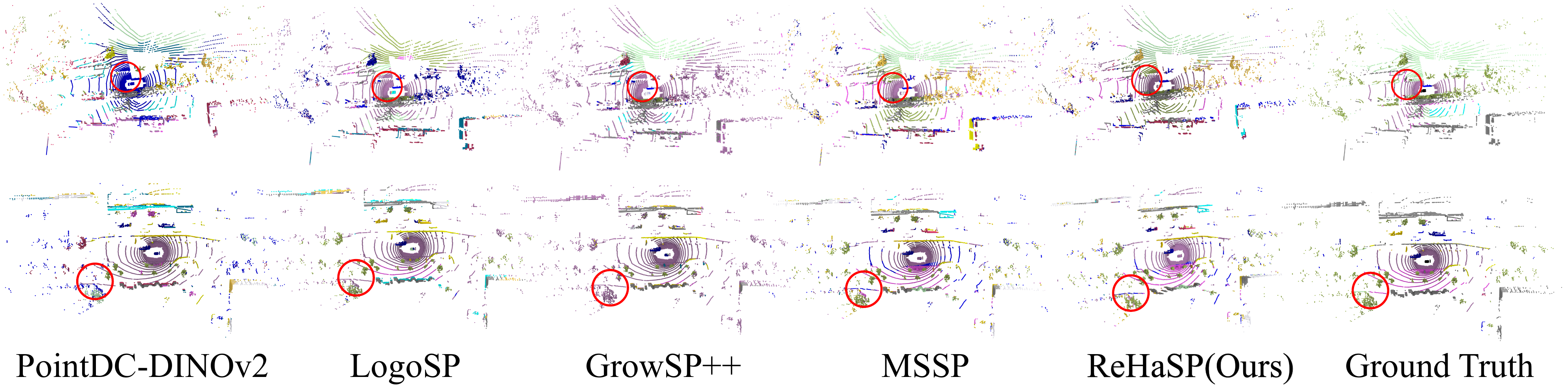}
	\caption{Qualitative comparison on nuScenes. Two outdoor scenes show the training-time priors on
		road-dominated scans.}
	\label{fig:qualnu}
\end{figure*}

\textbf{Results.}
As shown in Table~\ref{tab:nuscenes} and Fig.~\ref{fig:qualnu}, the training-time components remain positive, reaching
$\textbf{21.6}$ mIoU against LogoSP's $20.1$.
By contrast, SI \emph{reduces} mIoU by $0.3$ at $\gamma{=}0.4$.
We report this negative result rather than omitting it.

\textbf{Analysis.}
This is the failure mode derived in Section~\ref{sec:siboundary}. There the consensus is
road-dominated, so fusion moves prediction mass toward the majority classes and breaks more correct
labels than it cures, with $131{,}912$ flips against $69{,}013$ repairs. Superpoint size and purity
are not the operative moderator.
The $\gamma$ sweep confirms the mechanism, with losses deepening monotonically as points are pulled
harder toward the consensus (Fig.~\ref{fig:gamma}b).
A ten-seed replication (Fig.~\ref{fig:nuseed}) and the full-set GrowSP++ checkpoint
(Table~\ref{tab:universal}) show the deficit is systematic yet independent of training scale.
The training-time components are therefore robust across domains, and the sign of SI's gain is
governed by consensus reliability, not by training luck.

\subsection{Ablation study}
\label{sec:ablation}

Table~\ref{tab:ablation} reports the ablation study on S3DIS Area 5 and ScanNet.
Every component, granularity, and hyperparameter variant is evaluated under the same protocol,
and SI is applied only in the final row.

\begin{table}[t]
	\centering
	\caption{Results on nuScenes (OA/mAcc/mIoU, \%), evaluated on the full 6{,}018-scan val split; our
		models are trained on a 400-scan subset, ReHaSP uses $\gamma{=}0.4$. Best unsupervised results in
		\textbf{bold}.}
	\label{tab:nuscenes}
	\setlength{\tabcolsep}{6pt}
	\begin{tabular}{l ccc}
		\toprule
		Method & OA & mAcc & mIoU \\
		\midrule
		GrowSP \cite{Zhang_2023_CVPR}        & 39.2 & 17.5 & 10.2 \\
		PointDC \cite{Chen_2023_ICCV}        & 56.8 & 29.4 & 17.7 \\
		PointDC-DINOv2 \cite{Chen_2023_ICCV} & 51.8 & 28.6 & 18.2 \\
		LogoSP \cite{Zhang_2025_CVPR}        & 54.8 & 29.2 & 20.1 \\
		GrowSP++ \cite{zhang2026growsppp} $^\ddagger$ & \textbf{58.6} & \textbf{30.6} & 20.1 \\
		MSSP \cite{zhang2026mssp}                 & 55.5 & 30.1 & \textbf{21.6} \\
		\textbf{ReHaSP (Ours)}               & 55.4 & 29.6 & 21.3 \\
		\bottomrule
	\end{tabular}
	\par\vspace{3pt}\parbox{\linewidth}{\footnotesize
		$^\ddagger$Released checkpoint evaluated under the same protocol.}
\end{table}

\begin{table}[t]
	\centering
	\caption{Ablation study (mIoU, \%). (1)--(4) remove priors, (5)--(10) vary the scale set
		$\mathcal{K}$, (11)--(13) the temperature $\tau$, and (14)--(18) the merged-group count $C$, on
		both datasets; the defaults are $\tau{=}5$ and $C{=}40$ on ScanNet and $\tau{=}3$ and $C{=}80$ on
		S3DIS. All rows except (19), the full framework, are evaluated without SI, and row (3) quotes the
		published value. Best in each column in \textbf{bold}.}
	\label{tab:ablation}
	\setlength{\tabcolsep}{8pt}
	\begin{tabular}{l cc}
		\toprule
		Config & S3DIS Area 5 & ScanNet \\
		\midrule
		(1) w/o SC                & 45.3 & 30.3 \\
		(2) w/o MS                & 45.3 & 33.5 \\
		(3) w/o MS \& SC          & 46.5 & 35.8 \\
		\midrule
		(4) w/o SI                & 49.2 & 36.2 \\
		\midrule
		(5) $\mathcal{K}{=}\{80\}$      & 45.5 & 33.6 \\
		(6) $\mathcal{K}{=}\{40\}$      & 45.4 & 35.1 \\
		(7) $\mathcal{K}{=}\{20\}$      & 44.9 & 32.1 \\
		(8) $\mathcal{K}{=}\{80,40\}$   & 45.7 & 33.6 \\
		(9) $\mathcal{K}{=}\{40,20\}$   & 44.4 & 34.8 \\
		(10) $\mathcal{K}{=}\{80,20\}$  & 46.5 & 35.9 \\
		\midrule
		(11) $\tau{=}3$            & 49.2 & 34.7 \\
		(12) $\tau{=}5$            & 44.1 & 36.2 \\
		(13) $\tau{=}10$           & 45.5 & 32.9 \\
		\midrule
		(14) $C{=}20$              & 42.8 & 33.6 \\
		(15) $C{=}30$              & 41.5 & 31.8 \\
		(16) $C{=}40$              & 49.0 & 36.2 \\
		(17) $C{=}60$              & 43.9 & 34.1 \\
		(18) $C{=}80$              & 49.2 & 34.5 \\
		\midrule
		(19) The full framework    & \textbf{50.7} & \textbf{36.6} \\
		\bottomrule
	\end{tabular}
\end{table}

\emph{Component removal.}
Removing any prior degrades the framework on both datasets.
Row (1) drops to $45.3$ on S3DIS and collapses to $30.3$ on ScanNet, because multi-scale
descriptors amplify exactly the remote-fusion pathology when coarse-granularity features are
shared across disconnected regions, and the structurally repetitive ScanNet rooms suffer most.
Row (2) also fails on both datasets at $45.3$ and $33.5$, because enforcing spatial coherence on
the noisy single-scale space sacrifices feature-driven grouping.
Row (3), which matches the published setup of the strongest prior pipeline \cite{Zhang_2025_CVPR},
reaches $46.5$ and $35.8$.
Only \textbf{the full framework} recovers to $49.2$ and $36.2$ without SI, and SI lifts both to
\textbf{50.7} and \textbf{36.6} in row (19).
SI, by contrast, applies to any trained checkpoint without retraining, and its transfers to
other methods are positive on all five indoor baselines (Table~\ref{tab:universal}), suggesting a gain
orthogonal to the training-time components and in contrast with the nuScenes failure boundary
(Section~\ref{sec:siboundary}).

\emph{Granularity $\mathcal{K}$.}
The scale set $\mathcal{K}$ controls the granularities at which superpoint descriptors are
enriched before spectral analysis.
The default $\mathcal{K}{=}\{80,40,20\}$ pools at three granularities spanning object parts to
scene regions.
Rows (5)--(10) validate the choice by sweeping all proper subsets on both datasets.
Two patterns are consistent across the datasets.
First, the widest-span pair $\{80,20\}$ is the best subset on both, at $46.5$ on S3DIS and $35.9$
on ScanNet, so spanning the hierarchy matters more than including the object-level anchor.
Second, only the full tri-scale set clearly exceeds it, at $49.2$ and $36.2$ without SI and
$50.7$ and $36.6$ with SI.
The single scales diverge, S3DIS favoring the fine $\{80\}$ ($45.5$) and ScanNet the
object-level $\{40\}$ ($35.1$).
Every subset remains below its corresponding full configuration, echoing the coupling above.
These values were selected on ScanNet and transferred unchanged to S3DIS, so the choice of granularity involves no per-dataset tuning.

\emph{Temperature and consolidation.}
Rows (11)--(18) vary two further training-time hyperparameters on both datasets.
The temperature $\tau$ sharpens pseudo-label assignment.
On ScanNet the default is $\tau{=}5$ and both $\tau{=}3$ and $\tau{=}10$ lose accuracy around it.
On S3DIS the default is $\tau{=}3$, and $\tau{=}5$ and $\tau{=}10$ drop to $44.1$ and $45.5$.
High temperature disorders the clustering and collapses rare classes on both datasets.
The merged-group count $C$ controls the consolidation ratio.
On ScanNet, the default is $C{=}40$ and the curve is non-monotonic, peaking sharply at $40$ and
dropping to $33.6$ and $31.8$ at $20$ and $30$.
On S3DIS the default is $C{=}80$; intermediate consolidations drop sharply, to $42.8$, $41.5$, and
$43.9$ at $C{=}20$, $30$, and $60$, while $C{=}40$ stays competitive at $49.0$.
Notably, larger $C$ on ScanNet yields cleaner superpoint clustering yet no better final
segmentation, because finer merges mean finer-grained pseudo-labels with fewer points each and a
higher noise ratio.
We analyze the SI strength $\gamma$ in Section~\ref{sec:sianalysis}.

\subsection{Mechanism analysis: relocate then harvest}
\label{sec:mechanism}

\begin{figure*}[t]
	\centering
	\includegraphics[width=\textwidth]{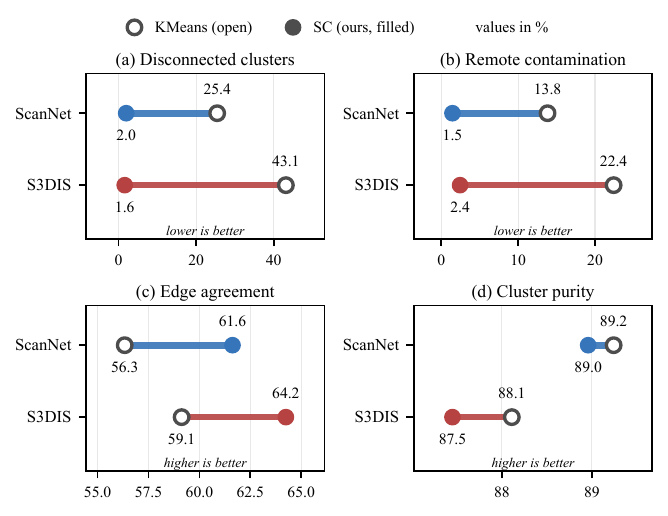}
	\caption{Structural statistics under KMeans and SC merging on ScanNet (100 scenes) and S3DIS
		(204). Dumbbells pair the merge rules (KMeans: gray open; SC: filled).
		(a)~Disconnected-cluster rate, (b)~remote cross-class contamination, (c)~edge
		agreement, (d)~cluster purity: SC reshapes (a)--(c) and leaves (d) essentially unchanged.}
	\label{fig:mechanism}
\end{figure*}

\begin{figure*}[t]
	\centering
	\includegraphics[width=\textwidth]{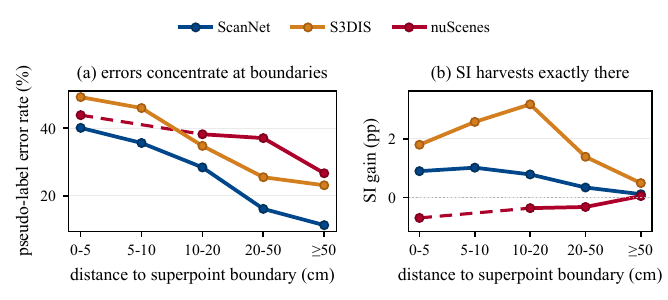}
\caption{Error structure and SI's gains by distance to the superpoint boundary. (a)~Pseudo-label
	error rate under SC merging; the profile is nearly identical under unconstrained merging.
	(b)~Per-point gain of SI on ScanNet, S3DIS, and nuScenes; on nuScenes ($0.15$\,m voxels; the
	empty 5--10\,cm bin shown as a dashed bridge) the profile inverts---losses concentrate at
	boundaries, where saturated agreement confidence fuses at near-full strength.}
	\label{fig:boundary}
\end{figure*}

Accuracy alone cannot distinguish ``purer pseudo-labels'' from ``differently structured
pseudo-labels''.
We therefore evaluate four label-free-to-compute quantities that use the ground truth only for
diagnosis.
The comparison is controlled for the merge rule: the same forward pass, the same spectral
primitives, and only
the merge rule toggled between unconstrained KMeans (the merge rule of the LogoSP pipeline)
and SC.
The isolation is possible because, in the MS configuration, the spectral path consumes
initial-superpoint features only.
The four quantities are:
(M1) \emph{pseudo-label purity}, the point-weighted fraction of points whose ground truth equals
the majority
ground truth of their pseudo-label group, at cluster and primitive granularity;
(M2) \emph{non-adjacent error fusion}, the rate of merged clusters with multiple connected
components, the fraction of cross-class superpoint pairs fused across components, and the point
contamination they cause;
(M3) \emph{graph-structure consistency}, the agreement between same-cluster and same-class decisions
over adjacency edges;
(M4) \emph{boundary errors}, error rates stratified by distance to the nearest superpoint boundary.
Stratifying errors by distance to a region boundary follows boundary-aware evaluation in
segmentation \cite{cheng2021boundary,kervadec2019boundary}. We adapt it to pseudo-labels, which
have no reference labels at training time.
Definitions are given in \ref{app:results}; the analysis below shows that they dissociate purity from
structure and localize where SI operates.
Fig.~\ref{fig:mechanism} visualizes the M1--M3 statistics for both merge rules, and every
quantity cited in this subsection is read off its panels.

\noindent\textbf{SC relocates errors (Fig.~\ref{fig:mechanism}).}
SC cuts the rate of disconnected clusters by $92\%$ and $96\%$ on ScanNet and S3DIS, remote cross-class
contamination by $89\%$ on both datasets, and improves graph-structure agreement by about
$+5$\,pp.
Yet cluster purity is essentially unchanged on both datasets.
Most remote fusions were dominated by a single class and therefore benign, and local merges slightly
increase adjacent cross-class pairs, offsetting the purification.
SC's contribution is thus \emph{structural}, not statistical: we use \emph{relocate} in the
budgetary sense---SC removes the errors scattered across remote fusions, and the residual,
already boundary-leaning under over-segmentation (Fig.~\ref{fig:boundary}a), becomes
boundary-dominated.
Fig.~\ref{fig:boundary}a quantifies the gradient: the error rate reaches $40$--$49\%$ within $5$\,cm
of a boundary, against $11$--$23\%$ at $50$\,cm and beyond, and Table~\ref{tab:boundarybins} reports
the complete per-bin distribution.
The gradient is nearly identical under unconstrained merging, with per-bin differences below
$0.8$\,pp on the indoor datasets, which confirms that SC relocates errors rather than changing
their per-bin rate.

The graph statistics of M3 further localize the relocation.
Over adjacency edges, SC \emph{trades keep precision for cut precision}: the probability that a
\emph{kept} edge is same-class drops slightly on both datasets, while the probability that a
\emph{cut} edge is indeed cross-class rises.

The relocation effect is likewise \emph{controlled for the feature scale}.
Because the merge operates on initial superpoints before any descriptor is formed, the M2 statistics
are bit-identical whether the downstream spectral path consumes single-scale or multi-scale
features; what changes is how the primitives absorb the relocated noise.
Under single-scale primitives, SC measurably lowers primitive purity, while under multi-scale
primitives, purity is nearly flat, as the enriched descriptors average out the boundary-local noise
that SC concentrates.
This is a second, independent reason for the regularize-then-enrich coupling observed in
Table~\ref{tab:ablation}: SC leaves a boundary-dominated residual noise field that MS's multi-scale
pooling, but not single-scale pooling, is robust to.

\noindent\textbf{SI harvests them (Fig.~\ref{fig:boundary}b).}
Stratifying SI's per-point corrections by the same boundary distance, gains concentrate exactly
where SC concentrates errors: $+0.9$--$+1.0$\, pp within 10\, cm on ScanNet against an overall
$+0.7$\,pp, and $+1.8$--$+3.2$\,pp on S3DIS with the peak at 10--20\,cm against an overall
$+1.9$\,pp, decaying $4$--$8\times$ toward region interiors.
The S3DIS peak sits farther from boundaries than ScanNet's because S3DIS scenes are larger and more
open, which widens the boundary band.
On the indoor pair, this offset tracks superpoint geometry rather than dataset identity.
The concentration itself depends on SC training.
Repeating the same stratification on MS-only checkpoints trained without SC (row~(1) of
Table~\ref{tab:ablation}) flattens or inverts the profile: ScanNet's ratio drops from $8.2\times$ to
$1.1\times$ (per-bin gains $+0.9$ to $+1.1$\,pp), and S3DIS's from $3.7\times$ to $0.4\times$, with the
boundary band now holding the \emph{smallest} gain (Table~\ref{tab:boundarybins}).
SI does not need SC to be positive; it needs SC to decide \emph{where} it is positive---the
harvest follows the relocation.
This control compares independently trained checkpoints rather than toggling a single model,
a weaker design than the merge-rule isolation above.

\noindent\textbf{Why coupling, mechanistically.}
The analysis also explains the ablation.
MS without SC enriches descriptors whose coarse granularities are exactly the ones involved in
remote fusion: enrichment without relocation amplifies the pathology.
SC without MS relocates errors of a representation too poor to benefit.
The two form one training-time mechanism: relocation without enrichment does not help on its
own, and enrichment without relocation is harmful.

\subsection{Analysis of SI: design, sensitivity, and applicability boundary}
\label{sec:sianalysis}
\label{sec:siboundary}

\noindent\textbf{Design space (Fig.~\ref{fig:gamma}a).}
Naive broadcasting, i.e., replacing every per-point prediction with its superpoint's,
\emph{hurts} on ScanNet, dropping from $35.7$ to $35.6$: mixed superpoints crush their minorities.
Hard gating, which broadcasts only for superpoints whose $\kappa_s$ exceeds a threshold, recovers
$+0.7$ on ScanNet; soft fusion of Eq.~\ref{eq:fusion} reaches $+0.9$ on the same checkpoint and
dominates on both indoor datasets across checkpoints.
Interestingly, naive broadcasting helps on S3DIS, $50.6$ against $49.2$ without voting, but hurts
on ScanNet and is catastrophic on nuScenes ($-2.3$\,pp).
Confidence-weighted soft fusion, by contrast, is positive on both indoor datasets and consistently
the best of the three designs; on nuScenes, where the consensus itself is wrong, even soft fusion
is slightly negative.
The comparison isolates \emph{confidence weighting}, not broadcasting itself, as the operative
ingredient.

\noindent\textbf{Per-class profile.}
Per-class gains are broad rather than class-specific, consistent with a boundary mechanism rather
than a class prior (\ref{app:perclass}).

\noindent\textbf{$\gamma$ sensitivity (Fig.~\ref{fig:gamma}b).}
S3DIS is flat over $\gamma\in[0.5,1.2]$, holding $50.5$ to $50.7$ with the peak at $0.8$, and
ScanNet is flat over $[0.3,1.2]$, holding $36.2$ to $36.6$ with the peak at $0.4$.
Only $\gamma\geq2$ collapses performance, to $35.4$ at $\gamma{=}2$ and $29.6$ at $\gamma{=}4$ on
ScanNet, when fusion overrides per-point evidence everywhere.
SI is thus insensitive to its single hyperparameter within each dataset's scale.
nuScenes, by contrast, loses \emph{monotonically} with $\gamma$, from $21.44$ at $\gamma{=}0.3$
through $21.30$ at $\gamma{=}0.4$ to $21.14$ at $\gamma{=}0.5$, a third regime in which there is no
benign operating point, consistent with the consensus-reliability boundary characterized below.

\noindent\textbf{SI transfers across methods (Table~\ref{tab:universal}, \ref{app:transfer}).}
Because SI only requires per-point predictions and superpoint partitions, it is mechanically
applicable to any superpoint-based unsupervised segmenter.
Applied without retraining to our reproduced LogoSP and PointDC-DINOv2 baselines and to the released GrowSP++
checkpoints, soft fusion is positive on all five combinations of method and dataset with strong
indoor baselines, by
$+0.1$ to $+2.2$ mIoU, using the same dataset-dependent $\gamma$ as our framework.
On the released nuScenes checkpoint, soft fusion loses $0.2$ mIoU, the outdoor
failure mode analyzed below.

\noindent\textbf{Confidence is label-free.}
The agreement confidence $\kappa_s$ of Eq.~\ref{eq:conf} correlates with true superpoint purity yet
requires no ground truth; homogeneous superpoints receive strong fusion, and mixed ones, typically
at object boundaries, remain per-point.
The confidence is also sharply \emph{bimodal}, which is what makes it act as a switch rather than a
blur (Fig.~\ref{fig:conf}): most points sit at $\kappa_s{>}0.9$, where the superpoint consensus
applies at full strength, while a substantial mass stays below $0.5$, where per-point evidence is
preserved, concentrated along superpoint boundaries---precisely the region SC's relocation fills
with errors.

\noindent\textbf{Applicability boundary, quantified.}
A natural hypothesis is that SI's gain tracks superpoint purity, but the hypothesis is ill-posed,
because purity depends on how it is counted.
Under the nominal, superpoint-averaged measure, nuScenes is the purest of the three datasets at
$0.997$, versus $0.91$ indoors, itself an artifact of size, since a ${\sim}57$-point outdoor
superpoint is almost always majority-dominated.
Under the point-weighted measure it is conversely the \emph{least} pure at $0.83$, versus
$0.92$--$0.94$ indoors.
Neither measure qualifies as an operative moderator, and what separates the datasets instead is
\emph{superpoint-consensus reliability}, which two accuracy measurements read off directly.
Indoors, the consensus is markedly more reliable than per-point prediction, at $93.7\%$ and
$97.3\%$ on ScanNet and S3DIS against per-point $72.9\%$ and $80.4\%$, so fusion is a denoising
operation.
On nuScenes, the consensus still wins scene-level accuracy while SI loses $0.3$ mIoU;
measured on the full validation split, the dissociation is a class-competition effect.
Scene-level pairing over all $6{,}018$ scans shows per-scene accuracy rising slightly ($+0.12$\,pp,
bootstrap $95\%$ CI $[0.07, 0.16]$) while mIoU falls $-0.30$: the consensus is road-dominated, so
fusion moves prediction mass toward the majority classes.
The classes that pay are exactly those whose per-point evidence beats the consensus (consensus vs.\
per-point accuracy: car $0.71$ vs.\ $0.82$, barrier $0.31$ vs.\ $0.46$, terrain $0.25$ vs.\
$0.37$, person $0.30$ vs.\ $0.36$), and among the $4.1\%$ of points whose predictions flip,
twice as many correct labels are broken as wrong ones cured ($131{,}912$ vs.\ $69{,}013$),
concentrated in person, terrain, and barrier.
The failure mode is thus not an unmeasurable ``domain gap'' but a measurable signature:
the per-class consensus advantage. Our diagnosis quantifies it against ground truth; a label-free
surrogate from prediction agreement alone is future work (Section~\ref{sec:conclusion}).
A ten-seed replication on nuScenes (Fig.~\ref{fig:nuseed}) confirms
that the outdoor deficit is systematic rather than a checkpoint artifact: the mean paired
$\Delta$ is $-0.21$\,pp (s.d.\ $0.12$; paired $t$-test $p{<}10^{-3}$, Wilcoxon $p{=}0.004$),
negative on nine of ten seeds, with worst $-0.34$ and best $+0.02$.
In deployment terms, SI should be switched on when the consensus is reliable; certifying that
reliability without labels remains open (Section~\ref{sec:conclusion}).

\subsection{Cross-dataset generalization}
\label{sec:crossdataset}

Table~\ref{tab:cross} in \ref{app:results} evaluates transfer in both directions without any fine-tuning.
ReHaSP denotes the present framework, i.e., our conference version MSSP \cite{zhang2026mssp} plus the
inference-time SI module.
Two findings emerge.

First, the framework generalizes well beyond the strongest published cross-dataset numbers in both
directions: $18.2$ against LogoSP's $16.9$ on S3DIS$\to$ScanNet ($+1.3$), and $48.5$
against $43.6$ on ScanNet$\to$S3DIS ($+4.9$).
Spatially coherent, multi-scale primitives survive the domain gap.
Second, \emph{the inference-time prior transfers across domains}: SI is positive in \emph{all twelve}
area and fold cells of Table~\ref{tab:cross}, by $+1.5$ to $+1.9$ per area on ScanNet$\to$S3DIS and
$+0.3$ to $+3.0$ per fold on S3DIS$\to$ScanNet, adding $+1.7$ and $+1.1$ mIoU on average on top of
the frozen conference backbone, the same direction-independent benefit it shows on the source
datasets, because cross-domain indoor models remain strong enough for superpoint consensus to be
reliable, in contrast to outdoor nuScenes where it is not, per the boundary analysis of
Section~\ref{sec:siboundary}.
This is \emph{not} domain adaptation \cite{wilson2020surveyuda}: no target
data enters training, so the protocol is strictly harder.

\section{Conclusion}
\label{sec:conclusion}
\sloppy
In this paper, we presented ReHaSP, a framework that makes both training and inference
structurally aware.
The training-time priors, SC and MS, act
as one mechanism---either alone does not help---and the inference-time SI
fuses per-point with superpoint-level predictions through a label-free agreement confidence at no
training cost.
A four-part mechanistic protocol showed that training moves the residual errors away from
spatially remote fusions, leaving them concentrated at superpoint boundaries where
inference harvests them---a relocate-then-harvest pattern.
Extensive experiments on three benchmarks and bidirectional cross-dataset transfers show that it outperforms all published unsupervised methods in mIoU on S3DIS Area 5 and the official ScanNet online benchmark, and, to our knowledge, is the first unsupervised result to surpass supervised PointNet++ on S3DIS Area 5.
We consider explicit analysis of error structure a complementary evaluation axis for pseudo-label
pipelines.

The framework has clear limits, each of which marks a direction for future work.
Primitive discovery remains class-imbalanced, and dominant primitives still absorb rare classes.
The outdoor failure boundary in Section~\ref{sec:siboundary} excludes autonomous-driving scenes from the benefits of SI until superpoint consensus becomes reliable.
SI is evaluated only as a training-free addition, and the mechanistic protocol still requires ground-truth labels for
diagnosis rather than for training.
Future work will integrate the fused predictions back into pseudo-label generation, seek a
label-free surrogate for the mechanistic metrics, and verify the pattern wherever pseudo-labels
inherit structure, such as 2D deep spectral segmentation \cite{melaskyriazi2022deepspectral} and
open-vocabulary 3D pipelines \cite{takmaz2023openmask3d}.
It will also ask whether the merge rule itself can adapt, so that adjacency tolerates the sparse,
occluded superpoints that mark the current outdoor boundary.



\section*{CRediT authorship contribution statement}
\textbf{Zhenghao Zhang:} Conceptualization, Methodology, Software, Investigation, Validation,
Writing -- original draft, Visualization.
\textbf{Xinjie Wang:} Methodology, Visualization, Funding acquisition, Writing -- review \& editing.
\textbf{Yifan Zhang:} Writing -- review \& editing.
\textbf{Wei Wang:} Conceptualization, Supervision.
\textbf{Jun Zhang:} Conceptualization, Supervision.
\textbf{Hanyun Wang:} Conceptualization, Resources, Supervision.

\section*{Declaration of competing interest}
The authors declare that they have no known competing financial interests or personal relationships
that could have appeared to influence the work reported in this paper.

\section*{Data availability}
ScanNet, S3DIS, and nuScenes are publicly available benchmarks. The code and trained checkpoints
will be released upon acceptance.

\section*{Funding}
This work was supported by the National Natural Science Foundation of China [Grant No. 62602685].

\section*{Declaration of generative AI and AI-assisted technologies in the manuscript preparation process}
During the preparation of this work the authors used Claude (Anthropic) in order to check
grammar and spelling and assist with language editing and plotting scripts.
After using this tool, the authors reviewed and edited the content as needed and take full
responsibility for the content of the published article.

\bibliographystyle{elsarticle-num}
\begingroup
\singlespacing
\setlength{\bibsep}{1pt plus 0.3ex}
\bibliography{references}
\endgroup

\clearpage
\appendix
\setcounter{topnumber}{4}
\setcounter{bottomnumber}{2}
\setcounter{totalnumber}{6}
\setcounter{figure}{0}
\setcounter{table}{0}
\section{Additional experimental results}
\label{app:results}
\sloppy

\begin{figure}[ht]
\centering
\includegraphics[width=0.95\textwidth]{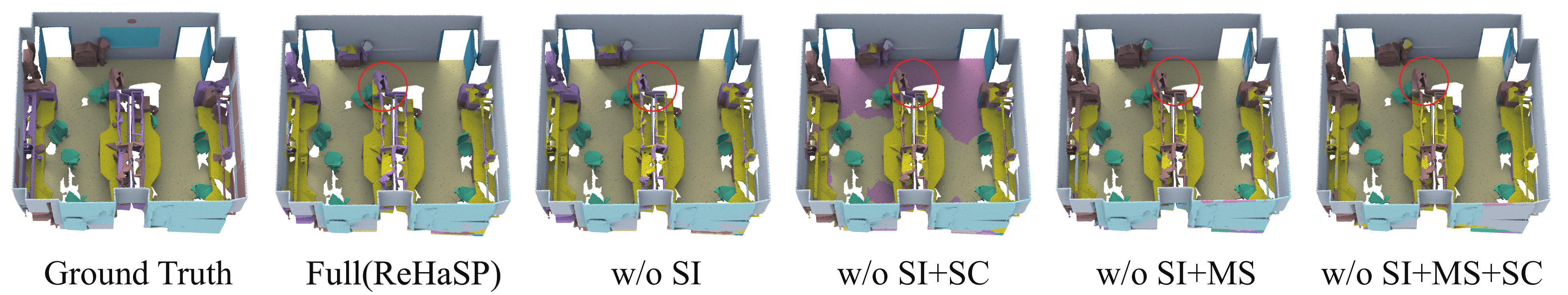}
\caption{Qualitative ablation on S3DIS Area 5; the full configuration (rightmost) removes the
residual boundary noise of its training-only variant.}
\label{fig:ablationvis}
\end{figure}

Let $H_i[a]$ count points of ground-truth class $a$ in initial superpoint $i$, $c(\cdot)$ the merge
assignment, and $\bar g(c)$ the majority class of cluster $c$.
\textbf{M1} cluster purity $=\sum_c \max_a \sum_{i\in c}H_i[a]$ $/$ $\sum_{i,a}H_i[a]$ (analogously
at primitive granularity).
\textbf{M2} (a) fraction of clusters whose induced adjacency subgraph has $>1$ connected component;
(b) fraction of cross-class same-cluster pairs lying in different components; (c) fraction of points
in clusters containing such pairs.
\textbf{M3} over adjacency edges $(i,j)$: agreement between $\mathbb{1}[c(i){=}c(j)]$ and
$\mathbb{1}[\bar g_i{=}\bar g_j]$, plus keep/cut precisions.
\textbf{M4} pseudo-label errors in distance bins $[0,5)$, $[5,10)$, $[10,20)$, $[20,50)$,
$\geq50$\,cm to the
nearest boundary voxel (6-adjacency): per-bin error rate, per-bin point mass, and the share of all
errors falling in each bin.
\fussy
Table~\ref{tab:boundarybins} reports the complete M4 stratification behind
Fig.~\ref{fig:boundary}.
Fig.~\ref{fig:ablationvis} provides the qualitative counterpart: the full configuration removes the
residual boundary noise that its training-only variant leaves near superpoint borders.

\begin{table}[ht]
\centering
\caption{Full M4 boundary-error stratification (Section~\ref{sec:mechanism}). For each distance
bin to the nearest superpoint boundary: point mass, pseudo-label error rate under unconstrained
KMeans and under SC merging, and the per-point gain of SI. The error gradient is nearly
merge-rule-invariant on the indoor datasets, with per-bin differences of at most $0.8$\,pp (the nuScenes
rates, by contrast, shift by up to $3.2$\,pp under SC), while SI's gain decays by roughly
$4$--$8\times$ from the boundary band to region interiors. nuScenes is evaluated at $0.15$\,m
voxels (its 5--10\,cm bin is empty).
The last column repeats the SI gain on MS-only checkpoints trained without SC (ScanNet and S3DIS):
without SC the boundary concentration collapses. nuScenes has no w/o-SC column because SI is already
negative there under the full framework.}
\label{tab:boundarybins}
\setlength{\tabcolsep}{5pt}
\begin{tabular}{l l c c c c c}
\toprule
& & & \multicolumn{2}{c}{Error rate (\%)} & & \\
\cmidrule(lr){4-5}
Dataset & Distance (cm) & Points (\%) & KMeans & SC & SI gain (pp) & w/o SC \\
\midrule
ScanNet & 0--5   & 19.2 & 40.2 & 40.1 & +0.90 & +0.89 \\
        & 5--10  & 22.1 & 35.6 & 35.6 & +1.02 & +1.08 \\
        & 10--20 & 20.4 & 28.3 & 28.4 & +0.78 & +0.92 \\
        & 20--50 & 26.0 & 16.0 & 16.1 & +0.34 & +0.66 \\
        & ${\geq}50$  & 12.2 & 11.3 & 11.3 & +0.11 & +0.84 \\
\midrule
S3DIS   & 0--5   & 34.8 & 49.4 & 49.2 & +1.80 & +0.51 \\
        & 5--10  & 25.5 & 46.0 & 46.0 & +2.58 & +0.83 \\
        & 10--20 & 12.9 & 34.1 & 34.8 & +3.18 & +1.09 \\
        & 20--50 & 18.7 & 24.7 & 25.5 & +1.39 & +1.18 \\
        & ${\geq}50$  & 8.2  & 23.2 & 23.1 & +0.49 & +1.24 \\
\midrule
nuScenes & 0--5   & 1.2 & 40.7 & 43.9 & $-0.70$ & -- \\
        & 5--10  & 0.0 & --- & --- & --- \\
        & 10--20 & 0.8 & 36.6 & 38.2 & $-0.37$ & -- \\
        & 20--50 & 2.8 & 35.4 & 37.1 & $-0.33$ & -- \\
        & ${\geq}50$  & 95.2 & 26.8 & 26.7 & +0.05 & -- \\
\bottomrule
\end{tabular}
\end{table}

\subsection{Full hyperparameters}
\label{app:hyper}

\begin{table}[ht]
\centering
	\caption{Complete configuration.}
\label{tab:hyper}
\begin{tabular}{l ccc}
\toprule
Hyperparameter & ScanNet & S3DIS & nuScenes \\
\midrule
Voxel size (m)        & 0.05 & 0.05 & 0.15 \\
Input features        & RGB & RGB & xyz \\
Distill epochs        & 300 & 700 & 300 \\
Feature dim $D$        & 384 & 384 & 384 \\
Primitives $P$         & 20 & 12 & 16 \\
Scales $\mathcal{K}$   & $\{80,40,20\}$ & $\{80,40,20\}$ & $\{20,10,5\}$ \\
Merged groups $C$      & 40 & 80 & 20 \\
Spectral columns       & 50 & 50 & 50 \\
Energy ratio           & 0.99 & 0.99 & 0.99 \\
Spectral sample cap    & 30{,}000 & 30{,}000 & 30{,}000 \\
Seg epochs             & 200 & 100 & 50 \\
Pseudo-label refresh   & \multicolumn{3}{c}{none (single clustering pass)} \\
Seg.\ optimizer / lr   & \multicolumn{3}{c}{AdamW / 1e-4 (BN momentum 0.02)} \\
Distill optimizer      & \multicolumn{3}{c}{Adam / 1e-3 (nuScenes: released ckpt)} \\
LR schedule            & \multicolumn{3}{c}{PolyLR} \\
Batch (distill / seg)  & \multicolumn{3}{c}{8 / 10} \\
Seed                   & 2022 & 2022 & 2023 \\
Superpoints            & \multicolumn{2}{c}{VCCS + region growing} & RANSAC + DBSCAN \\
Temperature            & 5 & 3 & 3 \\
SI $\gamma$            & 0.4 & 0.8 & 0.4 \\
Eval protocol          & \multicolumn{3}{c}{fixed-epoch checkpoint evaluation} \\
\bottomrule
\end{tabular}
\end{table}

Further reproducibility details: training applies random rotation, translation, and scaling of
the coordinates; colors are normalized to $[-0.5, 0.5]$ and are the only input features indoors
(xyz on nuScenes, Table~\ref{tab:hyper}).
Initial superpoints are VCCS supervoxels merged with normal-based region growing (pclpy) on
ScanNet and S3DIS, and RANSAC plane removal followed by DBSCAN on nuScenes.
DINOv2 ViT-S/14 patch-14 features are bicubically upsampled to each frame's $240{\times}320$
resolution before 2D--3D projection, and a point seen by several frames receives the average of
its projected features.
The mechanistic statistics of Section~\ref{sec:mechanism} use 100 ScanNet and 204 S3DIS training
scenes; all methods in Table~\ref{tab:universal} are evaluated under one shared protocol per
dataset.

\subsection{Sensitivity to the primitive count}
\label{app:psweep}
The class count $P$ is the pipeline's only prior, and it is set to the number of evaluated
categories by convention, as in earlier unsupervised pipelines.
Table~\ref{tab:psweep} sweeps $P$ on S3DIS Area 5 under the full training-time configuration
(SC, MS, $\tau{=}3$, $C{=}80$), holding the protocol fixed and varying only $P$; the anchor is a
same-seed local rerun, so cells compare within the table rather than against
Table~\ref{tab:s3dis}.
Values below the evaluated class count are not representable in the pipeline: merging the $P$
primitive centers into the evaluated categories requires $P \ge 12$ here.
At the fixed evaluation epoch, performance is flat to slightly better across a $4\times$ range
of $P$ ($46.3$, $46.3$, $47.7$ mIoU at $P{=}12/24/48$), so the oracle choice is not a knife-edge.
Larger $P$ instead destabilizes the training dynamics: $P{=}48$ collapses mid-training ($32.2$
mIoU at epoch 50, driven by fragmentation of the ceiling class) and only partially recovers,
and $P{=}24$ degrades late ($40.1$ at epoch 100).
This echoes the consolidation sweep of Section~\ref{sec:ablation}: finer pseudo-labels cover
fewer points each and carry a higher noise ratio, and the effect compounds with training
length, which the fixed-epoch protocol avoids.

\begin{table}[ht]
\centering
\caption{Sensitivity to the primitive count $P$ on S3DIS Area 5 (mIoU, \%; single seed,
self-anchored protocol; no bolding, cells compare within the table). $P{<}12$ is structurally
infeasible: the Hungarian protocol requires $P$ at least the number of evaluated categories.}
\label{tab:psweep}
\begin{tabular}{l ccc}
\toprule
$P$ & epoch 10 & epoch 50 & epoch 100 \\
\midrule
12 ($1\times$ classes) & 46.3 & 45.5 & 45.1 \\
24 ($2\times$)         & 46.3 & 46.6 & 40.1 \\
48 ($4\times$)         & 47.7 & 32.2 & 36.0 \\
\bottomrule
\end{tabular}
\end{table}



\begin{figure}[ht]
\centering
\includegraphics[width=0.78\textwidth]{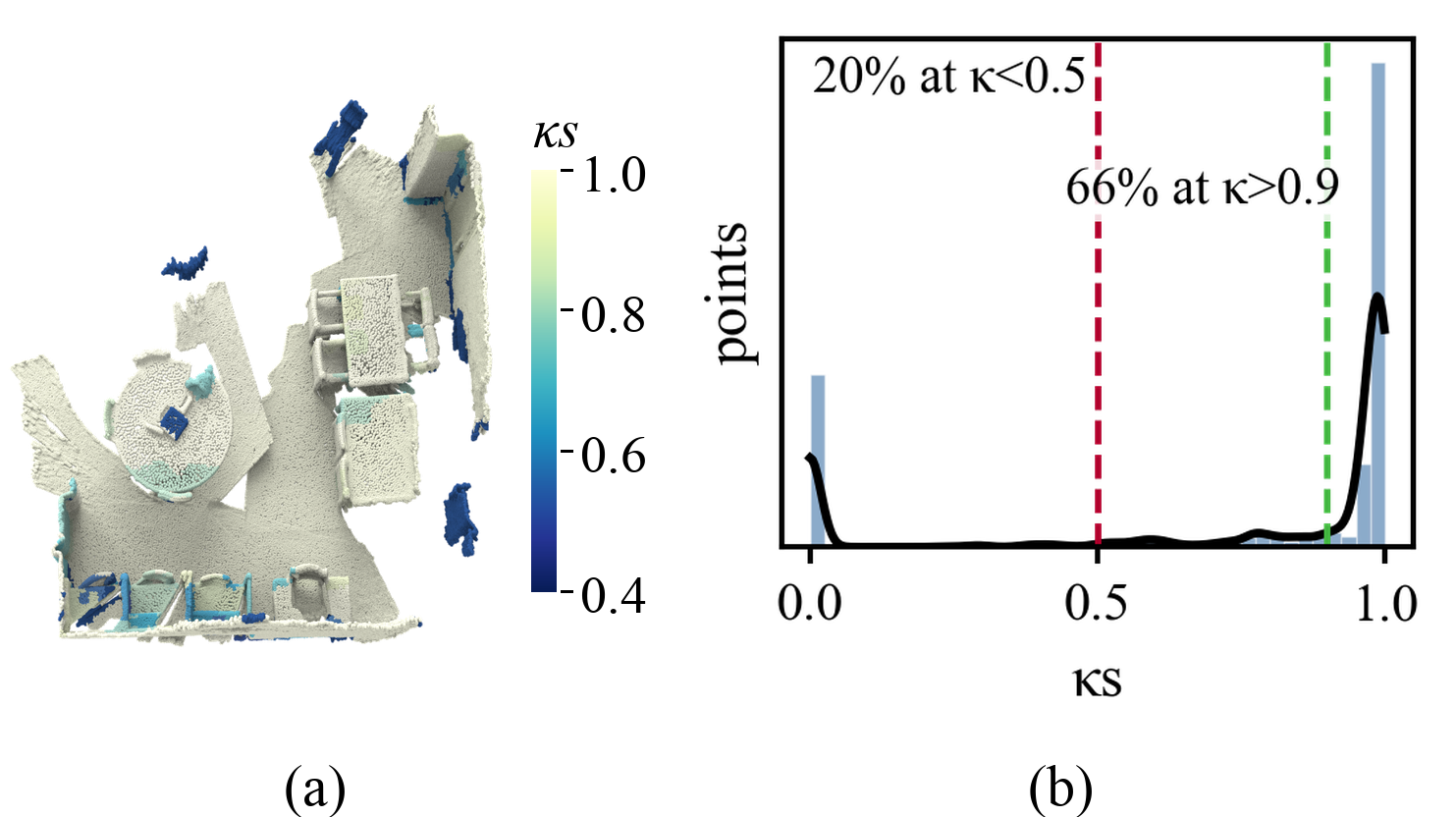}
\caption{(a)~Unsupervised agreement confidence $\kappa_s$ on a ScanNet scene, computed from
prediction agreement alone; bright (yellow, $\kappa_s{\to}1$) superpoints are homogeneous, dark ones are
mixed---typically at object boundaries, exactly where fusion must be gentle.
(b)~The distribution is bimodal, a switch: $66\%$ of points sit at $\kappa_s{>}0.9$ (full-strength
fusion) while $21\%$ stay below $0.5$ (per-point evidence kept).}
\label{fig:conf}
\end{figure}

Fig.~\ref{fig:conf} visualizes the agreement confidence $\kappa_s$ of Eq.~\ref{eq:conf} on ScanNet
validation scenes, supporting the bimodality discussion of Section~\ref{sec:sianalysis}.

\begin{figure}[ht]
\centering
\includegraphics[width=0.58\textwidth]{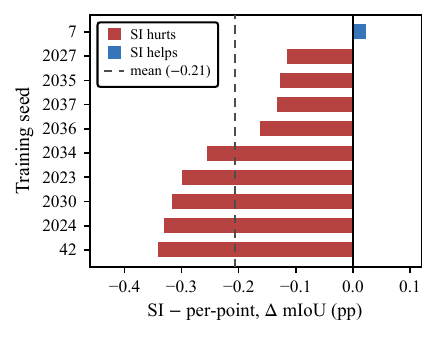}
\caption{nuScenes ten-seed replication of SI ($\gamma{=}0.4$): per-seed paired difference
in mIoU, SI minus per-point prediction (bars colored by sign; dashed line: mean). The mean
paired difference is $-0.21$\,pp (s.d.\ $0.12$, paired $t$-test $p{<}10^{-3}$, Wilcoxon
$p{=}0.004$; worst $-0.34$, positive on one of ten seeds). The outdoor deficit is systematic
yet bounded.}
\label{fig:nuseed}
\end{figure}

Fig.~\ref{fig:nuseed} visualizes the ten-seed nuScenes replication analyzed in
Section~\ref{sec:sianalysis}.

\begin{figure}[ht]
\centering
\includegraphics[width=0.98\textwidth]{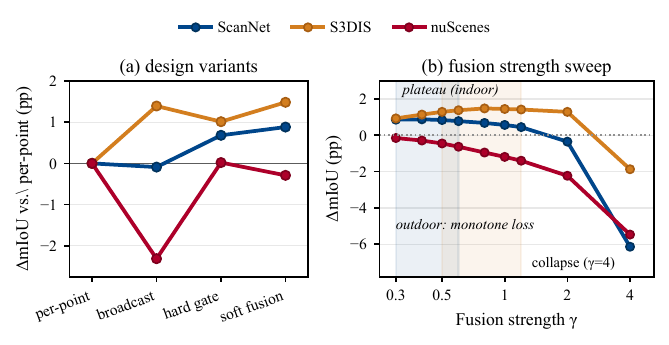}
\caption{Design variants and fusion strength of SI. (a)~mIoU (\%) of per-point inference,
broadcasting, hard gating, and soft fusion on the three datasets; broadcasting is
dataset-fragile---catastrophic on nuScenes---while confidence-weighted fusion is best on both
indoor datasets. (b)~Gain of soft fusion vs.\ fusion strength $\gamma$ on the three datasets
(log-$\gamma$ axis; the inset zooms into the indoor plateau, and the marked point is the ScanNet
collapse at $\gamma{=}4$).}
\label{fig:gamma}
\end{figure}

\subsection{Training-free SI transfer across methods}
\label{app:transfer}
Table~\ref{tab:universal} reports the training-free application of soft fusion to other methods'
checkpoints, analyzed in Section~\ref{sec:sianalysis}.
Its baseline rows are our reproductions or released official checkpoints evaluated under the same
protocol as Section~\ref{sec:setup}.

\begin{table}[ht]
\centering
\caption{Training-free application of soft SI fusion to other methods' checkpoints (our reproductions),
mIoU, \%; $\gamma{=}0.8$ on S3DIS, $0.4$ elsewhere; rows improved by SI in bold.}
\label{tab:universal}
\begin{tabular}{l ccc}
\toprule
Base method & w/o SI & $+$SI & $\Delta$ \\
\midrule
LogoSP, S3DIS Area 5        & 46.2 & \textbf{47.3} & $+1.1$ \\
LogoSP, ScanNet val         & 35.6 & \textbf{36.1} & $+0.5$ \\
PointDC-DINOv2, ScanNet val & 29.5 & \textbf{29.6} & $+0.1$ \\
GrowSP++$^\dagger$, ScanNet val & 33.9 & \textbf{34.5} & $+0.6$ \\
GrowSP++$^\dagger$, S3DIS Area 5 & 43.8 & \textbf{46.0} & $+2.2$ \\
GrowSP++$^\dagger$, nuScenes     & 20.1 & 19.9 & $-0.2$ \\
LogoSP, nuScenes             & 20.4 & 19.8 & $-0.6$ \\
\bottomrule
\end{tabular}
\par\vspace{3pt}\parbox{\linewidth}{\footnotesize
$^\dagger$Released official checkpoints evaluated under our unified protocol; base values are our
own evaluations of these checkpoints and can differ from the published numbers quoted in
Tables~\ref{tab:s3dis}--\ref{tab:nuscenes} (e.g., LogoSP on nuScenes 20.4 here vs.\ 20.1 published;
GrowSP++ on S3DIS 43.8 here vs.\ 46.6 published). PointDC-DINOv2 has no released S3DIS checkpoint.}
\end{table}

We also examined the most natural label-free surrogate for this reliability boundary: the macro
per-class consensus-to-point agreement, computable without any labels.
It takes its highest value exactly on nuScenes, $0.88$ against $0.77$ and $0.79$ on ScanNet and
S3DIS, where the road-saturated consensus is least reliable: agreement measures head similarity,
not consensus quality.
The label-free surrogate question therefore remains open (Section~\ref{sec:conclusion}).

\subsection{Computational cost}
\label{app:cost}
\begin{table}[ht]
\centering
\caption{Computational cost per component on ScanNet (RTX 4090).}
\label{tab:cost}
\setlength{\tabcolsep}{3pt}
\begin{tabular}{l cccc}
\toprule
Config & Train (s/ep) & Peak mem (GB) & Params & Infer. (ms/scene) \\
\midrule
LogoSP baseline & ${\sim}14$ & ${\sim}4.5$ & 30.6M & 13.4 \\
+SC             & adjacency$^*$ & negligible & 30.6M & 13.4 \\
+MS             & ${\sim}88$ (w/ SC) & 12.4 (spectral) & 30.6M & 13.4 \\
+SI             & 0 & $+$0 & $+$0 & $13.4{+}0.6$ \\
\bottomrule
\end{tabular}

\vspace{2pt}
{\footnotesize $^*$builds the per-scene adjacency graph, $O(N)$.}
\end{table}

Table~\ref{tab:cost} reports per-component cost, with all configurations measured on the same GPU
against our re-implemented LogoSP baseline, since cross-paper absolute times are not comparable
due to differing hardware and pipelines.
At inference, SI adds one scatter-mean pooling pass and a per-point fusion, $11$--$24$ MFLOPs per
scene; measured with a vectorized implementation it costs $0.6$\,ms per ScanNet scene ($4.5\%$ of
the $13.4$\,ms backbone forward) and $1.1$\,ms on nuScenes ($6.4\%$ of $17.3$\,ms), against the same
backbone as every baseline, with no parameters and no retraining.
The released evaluation scripts compute the same fusion in a per-superpoint Python loop for
clarity, which is slower but changes no prediction.
The baseline trains an epoch in ${\sim}14$\,s; the full MSSP configuration raises this to
${\sim}88$\,s, dominated by the $3\times$-expanded $1152$-dimensional superpoint descriptors and
the spectral pass.
SC's adjacency construction adds only $O(N)$ per scene; MS's dominant resource is the
${\sim}30\text{k}\times30\text{k}$ spectral decomposition at a 12.4\,GB peak, which caps the
superpoint sample budget.
In absolute terms, the complete 200-epoch segmentation stage finishes in under five hours on one
consumer GPU, excluding the one-off distillation stage.
At deployment, the framework is as light as its baseline: the parameter count is unchanged at
$30.6$\,M backbone parameters, backbone inference runs at $13.4$\,ms per ScanNet scene
(${\sim}75$ scenes/s) and $17.3$\,ms on nuScenes, and SI itself adds
no training cost, no parameters, and one superpoint pooling pass at $0.6$\,ms per scene.


\subsection{Full cross-dataset results}
\label{app:crossfull}
This subsection reproduces the complete per-area and per-fold cross-dataset table
behind Section~\ref{sec:crossdataset}.

\begin{table*}[ht]
\centering
\caption{Cross-dataset generalization without fine-tuning (mIoU, \%), transductive protocol with
$P{=}50$ primitives; $^{\dagger}$ReHaSP rows add measured SI increments to the published MSSP numbers
(re-trained folds use increments of the updated backbones). Best unsupervised in \textbf{bold}.
(a)~ScanNet $\to$ S3DIS, tested per area; (b)~S3DIS $\to$ ScanNet, trained on the areas listed and
tested on the held-out fold; means are simple averages.}
\label{tab:cross}
\textbf{(a)} ScanNet $\to$ S3DIS\par\smallskip
\setlength{\tabcolsep}{3pt}
\begin{tabular}{l ccccccc}
\toprule
test on $\to$ & Area 1 & Area 2 & Area 3 & Area 4 & Area 5 & Area 6 & Mean \\
\midrule
IIC \cite{ji2019iic}                  & 3.7 & 3.8 & 3.8 & 4.0 & 3.8 & 3.7 & 3.8 \\
PiCIE \cite{cho2021picie}              & 13.5 & 12.7 & 13.4 & 12.8 & 11.3 & 13.1 & 12.8 \\
GrowSP \cite{Zhang_2023_CVPR}          & 24.2 & 21.9 & 26.1 & 25.0 & 23.7 & 27.9 & 24.8 \\
PointDC \cite{Chen_2023_ICCV}          & 23.6 & 20.9 & 24.6 & 19.5 & 20.1 & 29.7 & 23.1 \\
PointDC-DINOv2 \cite{Chen_2023_ICCV}   & 33.8 & 29.6 & 33.8 & 31.7 & 32.2 & 36.9 & 33.0 \\
LogoSP \cite{Zhang_2025_CVPR}          & 43.8 & 37.5 & 47.0 & 40.7 & 44.9 & 47.9 & 43.6 \\
MSSP \cite{zhang2026mssp}              & 47.0 & 39.4 & 48.6 & 42.9 & 48.7 & 53.9 & 46.8 \\
\textbf{ReHaSP}$^{\dagger}$ & \textbf{48.9} & \textbf{41.3} & \textbf{50.4} & \textbf{44.7} & \textbf{50.2} & \textbf{55.4} & \textbf{48.5} \\
\bottomrule
\end{tabular}

\vspace{8pt}
\textbf{(b)} S3DIS $\to$ ScanNet (model trained on the five areas listed)\par\smallskip
\setlength{\tabcolsep}{3pt}
\begin{tabular}{l ccccccc}
\toprule
trained on $\to$ & {\footnotesize 2/3/4/5/6} & {\footnotesize 1/3/4/5/6} & {\footnotesize 1/2/4/5/6} & {\footnotesize 1/2/3/5/6} & {\footnotesize 1/2/3/4/6} & {\footnotesize 1/2/3/4/5} & Mean \\
\midrule
IIC \cite{ji2019iic}                  & 3.5 & 3.4 & 3.7 & 3.5 & 3.5 & 3.6 & 3.5 \\
PiCIE \cite{cho2021picie}              & 5.6 & 5.1 & 5.0 & 5.9 & 6.0 & 5.5 & 5.5 \\
GrowSP \cite{Zhang_2023_CVPR}          & 16.9 & 17.8 & 16.4 & 16.1 & 17.1 & 15.3 & 16.6 \\
PointDC \cite{Chen_2023_ICCV}          & 10.3 & 10.1 & 10.4 & 8.4 & 10.0 & 9.9 & 9.9 \\
PointDC-DINOv2 \cite{Chen_2023_ICCV}   & 16.7 & 15.4 & 16.4 & \textbf{17.6} & 14.3 & 14.6 & 15.8 \\
LogoSP \cite{Zhang_2025_CVPR}          & 16.9 & 16.9 & 16.8 & 16.8 & 16.7 & \textbf{17.0} & 16.9 \\
MSSP \cite{zhang2026mssp}              & 17.4 & 17.1 & 17.1 & 16.8 & 18.0 & 16.2 & 17.1 \\
\textbf{ReHaSP}$^{\dagger}$ & \textbf{17.7} & \textbf{20.1} & \textbf{18.3} & 17.4 & \textbf{18.5} & 16.9 & \textbf{18.2} \\
\bottomrule
\end{tabular}
\end{table*}

\subsection{Per-class results}
\label{app:perclass}
Tables~\ref{tab:perclass} and \ref{tab:perclasssc} report per-class IoU on S3DIS Area 5 and the
ScanNet validation split under the same Hungarian matchings as the main tables.
The per-class profile shows how the training-time gain arises on S3DIS: it concentrates in a few
large-structure classes, most visibly sofa ($+42.9$, from $21.9$ to $64.8$) and wall ($+3.3$),
while eight of the remaining eleven classes decline slightly; excluding sofa, the training-time
priors average $-0.95$ IoU across those classes.
The sofa jump reflects primitive realignment under a different Hungarian matching---sofa-like
regions consolidate into a coherent primitive---rather than a broad per-class improvement, and
ScanNet shows the same realignment character (wall and desk $+20.7$ each, chair $-27.1$,
other furniture $-13.4$).
SI, by contrast, adds small positive margins almost uniformly, consistent with its boundary
mechanism.

\clearpage
\begin{table}[ht]
\centering
\caption{Per-class IoU (\%) on S3DIS Area 5. $\Delta_{\mathrm{t}}$: gain of the training-time
priors (w/o SI over LogoSP); $\Delta_{\mathrm{SI}}$: further gain of SI. Best per row in
\textbf{bold}; the training-time gains concentrate in a few large-structure classes (see text).}
\label{tab:perclass}
\small
\begin{tabular}{l ccccc}
\toprule
Class & LogoSP & w/o SI & ReHaSP & $\Delta_{\mathrm{t}}$ & $\Delta_{\mathrm{SI}}$ \\
\midrule
ceiling & 93.9 & 92.9 & \textbf{94.2} & -1.0 & +1.3 \\
floor & \textbf{90.8} & 86.6 & 89.6 & -4.2 & +3.0 \\
wall & 68.1 & 71.4 & \textbf{76.0} & +3.3 & +4.6 \\
beam & \textbf{0.3} & 0.0 & 0.0 & -0.3 & +0.0 \\
column & \textbf{4.5} & 1.7 & 1.8 & -2.8 & +0.1 \\
window & 56.9 & 59.5 & \textbf{61.5} & +2.6 & +1.9 \\
door & \textbf{46.7} & 43.9 & 46.2 & -2.7 & +2.2 \\
chair & \textbf{60.9} & 53.2 & 54.2 & -7.7 & +1.0 \\
table & 43.0 & 47.1 & \textbf{48.3} & +4.1 & +1.2 \\
sofa & 21.9 & 64.8 & \textbf{66.7} & +42.9 & +1.8 \\
bookcase & \textbf{70.6} & 69.0 & 69.5 & -1.6 & +0.5 \\
board & \textbf{0.2} & 0.1 & 0.1 & -0.1 & +0.0 \\
\bottomrule
\end{tabular}
\end{table}

\begin{table}[ht]
\centering
\caption{Per-class IoU (\%) on the ScanNet validation split. $\Delta_{\mathrm{t}}$: gain of the
training-time priors (w/o SI over LogoSP); $\Delta_{\mathrm{SI}}$: further gain of SI. Best per row in
\textbf{bold}; per-class differences partly reflect the different Hungarian alignments.}
\label{tab:perclasssc}
\small
\begin{tabular}{l ccccc}
\toprule
Class & LogoSP & w/o SI & ReHaSP & $\Delta_{\mathrm{t}}$ & $\Delta_{\mathrm{SI}}$ \\
\midrule
wall & 46.3 & 67.0 & \textbf{68.4} & +20.7 & +1.4 \\
floor & 86.6 & \textbf{90.1} & \textbf{90.1} & +3.5 & +0.0 \\
cabinet & 20.7 & 32.0 & \textbf{32.7} & +11.3 & +0.8 \\
bed & \textbf{66.8} & 61.5 & 61.7 & -5.3 & +0.2 \\
chair & \textbf{63.3} & 36.2 & 36.3 & -27.1 & +0.1 \\
sofa & \textbf{50.9} & 49.7 & 50.4 & -1.2 & +0.7 \\
table & 47.1 & 48.5 & \textbf{49.3} & +1.4 & +0.7 \\
door & 33.8 & \textbf{37.1} & 35.8 & +3.3 & -1.3 \\
window & 41.6 & \textbf{42.2} & \textbf{42.2} & +0.6 & +0.0 \\
bookshelf & \textbf{62.8} & 57.5 & 58.2 & -5.3 & +0.7 \\
picture & \textbf{1.0} & 0.0 & 0.0 & -1.0 & -0.0 \\
counter & \textbf{38.0} & 36.7 & 37.6 & -1.3 & +0.8 \\
desk & 10.5 & 31.2 & \textbf{32.0} & +20.7 & +0.8 \\
curtain & \textbf{28.6} & 18.9 & 18.2 & -9.7 & -0.7 \\
refrigerator & \textbf{0.5} & 0.1 & 0.1 & -0.4 & +0.0 \\
shower curtain & 0.0 & \textbf{0.3} & 0.2 & +0.3 & -0.0 \\
toilet & \textbf{46.3} & 42.5 & 43.9 & -3.8 & +1.3 \\
sink & 0.0 & 0.0 & 0.0 & +0.0 & +0.0 \\
bathtub & 42.3 & 56.3 & \textbf{57.8} & +14.0 & +1.4 \\
other furniture & \textbf{29.6} & 16.2 & 16.3 & -13.4 & +0.1 \\
\bottomrule
\end{tabular}
\end{table}

\clearpage
\section{Complete algorithm}
\label{app:alg}
\renewcommand{\thealgorithm}{B.\arabic{algorithm}}
\begingroup\singlespacing
Algorithm~\ref{alg:framework} details the complete procedure of the framework:
Stage~1 cross-modal distillation (Section~\ref{sec:prelim}), Stage~2 structural primitive
discovery under the SC and MS priors (Section~\ref{sec:discovery}), and Stage~3 structure-aware
inference (Section~\ref{sec:si}).
Hyperparameters follow Table~\ref{tab:hyper}.

\begin{algorithm}[H]
\caption{The complete framework: structural priors across training and inference}
\label{alg:framework}
\begin{algorithmic}[1]
\REQUIRE scenes $\{\mathcal{P}^m\}$ with multi-view images; scales $\mathcal{K}$; merged groups $C$; primitives $P$; fusion strength $\gamma$
\ENSURE fused predictions $\{\tilde y_n\}$ for all points, without any labels
\STATE \textbf{Stage 1: 2D$\to$3D distillation} (Sec.~\ref{sec:prelim}): train backbone $f_\theta$ by $\mathcal{L}_{\text{distill}}$ (Eq.~\ref{eq:distill}) against projected DINOv2 features
\STATE \textbf{Stage 2: structural primitive discovery} (Sec.~\ref{sec:discovery})
\FOR{each scene $\mathcal{P}^m$}
  \STATE $h_s \leftarrow$ average-pool $f_\theta$ features within each initial superpoint $\mathcal{R}_s$ (Eq.~\ref{eq:spfeat})
  \STATE \textbf{SC (Sec.~\ref{sec:sc})}: build adjacency graph $\mathcal{G}^m$ (Eq.~\ref{eq:adj}); merge $\{\hat h_s\}$ into $C$ groups by \emph{connectivity-constrained} agglomeration
  \STATE \textbf{MS (Sec.~\ref{sec:ms})}: $g_s \leftarrow [\,g_s^{(K_1)};\ldots;g_s^{(K_{|\mathcal{K}|})}\,]$ by clustering $\{\hat h_s\}$ at each granularity $K\!\in\!\mathcal{K}$ (Eqs.~\ref{eq:ms_centroid})
\ENDFOR
\STATE spectral analysis on $\{g_s\}$: RBF affinity $\to$ Laplacian eigenvectors $\to$ KMeans $\to$ primitives $\{\phi_1,\ldots,\phi_P\}$ with centers $W$ in the $384$-d space
\STATE assign each merged group to its cosine-nearest primitive center; propagate to points as pseudo-labels
\STATE \textbf{Training on pseudo-labels}: freeze $W$; train $f_\theta$ by minimizing $\mathcal{L}_{\text{seg}}$ (Eq.~\ref{eq:ce}) on the pseudo-labels
\STATE \textbf{Stage 3: structure-aware inference} (Sec.~\ref{sec:si})
\FOR{each point $n$}
  \STATE $z_n \leftarrow \cos\bigl(f_\theta(x_n),\, W\bigr)$ \COMMENT{per-point logits}
\ENDFOR
\FOR{each superpoint $s$}
  \STATE $\bar z_s \leftarrow \mathrm{normalize}\!\bigl(\tfrac{1}{|\mathcal{R}_s|}\textstyle\sum_{n \in \mathcal{R}_s} f_\theta(x_n)\bigr) W$ \COMMENT{SI: consensus logits}
  \STATE $\kappa_s \leftarrow \tfrac{1}{|\mathcal{R}_s|}\sum_{n\in\mathcal{R}_s}\mathbb{1}\!\left[\arg\max z_n = \arg\max \bar z_s\right]$ \COMMENT{label-free agreement}
\ENDFOR
\FOR{each point $n$ in superpoint $s(n)$}
  \STATE $\tilde z_n \leftarrow z_n + \gamma\,\kappa_{s(n)}\,(\bar z_{s(n)} - z_n)$;\quad $\tilde y_n \leftarrow \arg\max \tilde z_n$ \COMMENT{SI: soft fusion}
\ENDFOR
\RETURN $\{\tilde y_n\}$
\end{algorithmic}
\end{algorithm}
\endgroup

\end{document}